\documentclass[11pt]{article}
\usepackage{arxiv}

\usepackage[T1]{fontenc}

\usepackage{amsmath,amssymb,amsbsy,amsfonts}
\usepackage{graphicx}
\IfFileExists{subfigure.sty}{\usepackage{subfigure}}{}
\IfFileExists{epsfig.sty}{\usepackage{epsfig}}{}
\usepackage{multirow}
\usepackage{booktabs}
\usepackage{array}
\usepackage{comment}
\usepackage{xcolor}
\graphicspath{{./}}
\usepackage[hidelinks]{hyperref}
\usepackage{url}
\usepackage{natbib}

\newcolumntype{L}[1]{>{\raggedright\arraybackslash}p{#1}}
\newcolumntype{C}[1]{>{\centering\arraybackslash}p{#1}}

\setcitestyle{authoryear,round,aysep={,},yysep={;}}

\newcommand{\ECHead}[1]{\clearpage\begin{center}\Large\bfseries #1\end{center}\vskip 1em}
\newcommand{\ECSwitch}{%
  \appendix
  \setcounter{section}{0}%
  \renewcommand{\thesection}{EC.\arabic{section}}%
  \renewcommand{\thesubsection}{EC.\arabic{section}.\arabic{subsection}}%
  \setcounter{equation}{0}\renewcommand{\theequation}{EC.\arabic{equation}}%
  \setcounter{table}{0}\renewcommand{\thetable}{EC.\arabic{table}}%
  \setcounter{figure}{0}\renewcommand{\thefigure}{EC.\arabic{figure}}%
}

\title{When Should Service Agents Reconsider? Difficulty-Routed Control in Customer-Service Operations\thanks{All authors contributed equally and are listed alphabetically by last name.}}

\author{%
  {\large Qian Chen\textsuperscript{$\dagger$}\qquad Chengyuan Liu\textsuperscript{$\ddagger$}\qquad Xin Yu\textsuperscript{$\ddagger$}}\\[0.8em]
  \textsuperscript{$\dagger$}Department of Supply Chain and Information Systems\\[0.15em]
  \textsuperscript{$\ddagger$}Department of Statistics\\[0.15em]
  The Pennsylvania State University\\[0.5em]
  \texttt{\{quc20,\;cjl6934,\;xmy5152\}@psu.edu}\\[0.7em]
}

\begin{document}
\maketitle

\begin{abstract}
Autonomous customer-service agents are shifting from conversational interfaces toward operational execution roles: they retrieve firm records, apply service policies, and execute backend writes such as refunds, cancellations, exchanges, order modifications, and reservation changes. This shift creates a service-control problem: firms must keep routine service fast and low-friction while preventing operational errors on requests where customer instructions, policy constraints, firm records, and backend writes interact. We propose a difficulty-routed service-control architecture that asks when service agents should reconsider before acting. A lightweight router keeps routine sessions on a low-cost baseline path and routes operationally coupled sessions to an escalated workflow. The escalated path uses conflict-aware communication and write-triggered reconsideration to concentrate deliberation and safeguards before consequential backend writes, rather than applying additional control uniformly across all service sessions. We evaluate the architecture on human-verified retail and airline tasks from $\tau^{2}$-bench. In retail, the method improves reliability consistently on service requests with operational conflict. Routing evidence shows that stronger control is directed toward conflicted requests rather than broadly applied to routine ones. Dialogue and tool-use profiles suggest that gains do not come from indiscriminate interaction expansion or broader tool chains; instead, added turns and tool calls support evidence gathering, write separation, and pre-write reconsideration. Case-level evidence shows that the escalated workflow preserves fallback plans, binds retrieved records to the correct action, sequences writes, and decomposes multi-entity requests. Airline results extend the same service-control logic to reservation operations. The paper contributes a service-control perspective on agentic AI by showing how routing and write-triggered reconsideration can selectively allocate communication, deliberation, and safeguards across heterogeneous customer-service requests.
\end{abstract}


\vskip 1em

\section{Introduction}
\label{section:introduction}

Generative AI is changing customer-service operations from a communication technology into an execution technology. Early service chatbots primarily answered questions, provided policy information, or recommended next steps. Newer autonomous service agents can converse with customers, retrieve account or order information, apply business rules, and execute \emph{backend writes}---state-changing updates to firm records, such as returns, exchanges, cancellations, address changes, refunds, and reservation updates \citep{huang2024caring,wang2026agentic}.
This shift creates opportunities for faster and more scalable service delivery, but it also changes the nature of operational risk. When an AI system can directly modify service records, an error is no longer merely an incorrect response; it can become an incorrect refund, cancellation, exchange, booking change, or customer commitment.

This operational risk is difficult to manage because customer-service requests are heterogeneous. Many requests are routine: a customer asks for a straightforward return, a simple cancellation, a policy explanation, or the status of an order. These cases should be handled quickly, at low cost, and with minimal customer friction. Other requests create \textit{operational conflict}: the correct action depends on how customer instructions, firm records, policy constraints, and backend writes interact. Some requests require the agent to preserve conditional or fallback instructions, such as ``exchange it if the requested item is available, otherwise refund me.'' Others become difficult because the customer revises a pending action before execution, such as changing the requested item, payment method, or cancellation scope after the agent has proposed an action. Some requests require the agent to bind retrieved records to the correct write target: the relevant order, item, address, reservation, passenger, or payment method may only become clear after record retrieval. Finally, some requests require the agent to preserve the correct scope and sequence of backend writes. A cancellation may apply to one item rather than an entire order, or an early write may change the state and make a later required update infeasible. In these cases, the agent must not only understand the customer's request, but also decide what to write, to which object, with what scope, and in what order.

This creates a service-control tradeoff. Treating all requests as complex would increase computation, latency, and unnecessary customer-facing confirmation burden. Treating all requests as routine would expose the firm to avoidable service failures when the agent must coordinate multiple goals, records, or backend writes. The central design question is therefore not simply whether an autonomous service agent can complete a task, but how a service system should allocate stronger control across heterogeneous customer-service requests.

Prior research motivates the study of autonomous AI in customer service, but it does not directly address the control-allocation problem created by autonomous service execution. Marketing and service-operations work has examined how customers respond to AI in service encounters and how generative AI can improve service productivity \citep{huang2018artificial,puntoni2021consumers,huang2024caring,wang2023voice,brynjolfsson2025generative}. More recent work has begun to study agentic AI systems that autonomously perform customer-service tasks, often with human-in-the-loop interventions to manage escalations and recover from AI failures \citep{wang2026agentic}. In parallel, the agent-systems literature has developed benchmarks and verification methods for tool-using agents that must follow policies and update backend systems \citep{yao2024tau,barres2025tau,kamath2025enforcing}. Much of this work evaluates AI systems as customer-facing interfaces, productivity tools, or agents to be improved on aggregate benchmark performance.Less attention has been paid to when an autonomous service agent should proceed routinely, when it should clarify customer instructions, and when it should reconsider a backend write before committing it.

We study this problem through the lens of \textit{difficulty-routed service control}. Our premise is that deliberation, customer-facing clarification, and safeguards are operational resources: they can prevent failures in complex requests, but they also consume computation, increase latency, and may add customer-facing turns. Rather than applying the same workflow to all interactions, we ask when an autonomous service agent should handle a request on the baseline path, using the standard workflow, and when it should escalate to stronger control before executing backend writes. The design objective is therefore not simply to make the agent reason more, but to allocate stronger control to interactions where operational conflict makes errors more likely or more costly.

We operationalize this idea through a \emph{difficulty-routed service-control} architecture with two linked modules. First, a \textit{difficulty router} determines whether an ongoing service session should remain on the baseline path or be escalated to a higher-control path. The router is designed to detect \emph{operational coupling} rather than linguistic complexity. Operational coupling arises when the agent must coordinate multiple requested service actions, reason across multiple operational entities such as orders, items, reservations, passengers, or payment methods, preserve conditional or fallback instructions, resolve conflicting customer constraints, or execute state-changing actions in an order that affects feasibility. Second, once a session enters the escalated path, an \textit{escalated workflow} adds control at two operationally important points: when the agent must resolve ambiguous or conflicting customer instructions, and when it is about to execute a state-changing backend action. The workflow first uses a Reason-Speak-Act (ReSpAct) prompt \citep{yao2022react,dongre2025respact} to generate the next candidate action, which can be a customer-facing message, a read-only lookup of firm records, or a state-changing backend write. This makes clarification or conflict communication part of the agent's action space when customer instructions, firm records, or policies create competing interpretations. Before state-changing backend writes, the workflow then applies write-triggered reconsideration through a pre-write verifier, which checks whether the proposed action targets the correct record, uses the correct scope and arguments, preserves unresolved customer constraints, and avoids blocking later required actions. Together, these modules allow the agent to preserve efficiency for routine requests while concentrating additional control on interactions where service failures are more likely or harder to reverse.

We evaluate the architecture using human-verified customer-service tasks from $\tau^{2}$-bench \citep{barres2025tau}. Retail serves as the primary empirical setting because it contains a rich concentration of operational conflict, including fallback requests, confirmation-triggered revisions, cross-order dependencies, multi-write execution, payment constraints, and inventory-driven substitutions. These features make retail especially useful for studying when stronger service control should be allocated and how pre-write safeguards affect outcomes. We use airline as a structurally distinct secondary service domain to examine whether the same control logic extends beyond retail to reservation-service operations, where operational conflict arises through reservation sequencing, multi-segment updates, travel certificates, payment ordering, and policy-gated eligibility constraints.

Because stronger control is not expected to help every task, we construct an \textit{evaluation focus set} in each domain. The focus set identifies tasks whose baseline transcripts exhibit operational conflict, such as multiple executable writes, incompatible intents, confirmation-triggered revisions, late-emerging constraints, or conditional fallback plans. This design lets us evaluate not only aggregate task success, but also whether stronger control improves reliability on the service requests where control is theoretically most warranted.

The results support the value of the architecture as a targeted service-control policy. In the primary retail setting, the routed architecture improves majority-pass performance on the evaluation focus set across all three retail configurations, precisely where operational conflict makes stronger control most relevant. By contrast, aggregate full-task-set effects are more mixed, which is consistent with the design objective: the architecture is not intended to add control uniformly to every request, but to improve reliability where routine execution is most likely to be insufficient. 

The routing analysis shows that the architecture achieves this selective allocation. Escalated-path activation is concentrated on tasks with operational conflict rather than spread broadly across routine requests. In the primary retail configuration, every task routed to the escalated path belongs to the evaluation focus set, indicating that the router is not simply escalating busy or lengthy conversations. Instead, it identifies requests where customer instructions, firm records, policy constraints, and backend writes must be coordinated before action. Routing analysis further shows that escalated-path activation captures both \emph{immediate conflict}, where the customer's first request already contains multiple actions, fallback conditions, or scope constraints, and, more often, \emph{emergent conflict}, where an initially routine request later reveals a control problem through retrieved records, policy or availability constraints, or customer revisions before a write.

The stable-gain case analysis further shows that the gains are not driven by indiscriminate dialogue expansion or tool use. Dialogue and tool-use profiles suggest that the escalated path does not simply add more interaction or broader tool chains. Rather, additional turns cluster around risky write decisions, while tool calls reflect repeated use of core tools for evidence gathering, write separation, and local checking. Case-level evidence clarifies how this targeted control improves outcomes: the escalated path preserves fallback plans, binds retrieved records to the correct backend action, sequences coupled writes, and decomposes multi-entity requests into locally checked write units. These patterns address the specific failure modes that arise when an agent writes to the wrong object, uses the wrong scope, or executes actions in the wrong sequence. Airline reservation operations provide a secondary-domain extension, showing analogous routing and recovery patterns in a structurally different setting where operational coupling arises through payment instruments, reservation structure, policy-gated changes, and coordinated reservation writes.

The paper makes three contributions. First, we frame autonomous service agents as a control problem at the operations--marketing interface. Existing work has studied AI service agents as customer interfaces, productivity tools, or benchmarked tool users. We instead emphasize the coordination between customer-facing interaction and backend service execution. Customer communication shapes service speed, friction, and customer understanding, while backend writes determine whether the final service outcome is operationally correct. This perspective shifts attention from whether an agent completes tasks in aggregate to when additional clarification, deliberation, and safeguards are needed to prevent operational errors before consequential service actions are committed.

Second, we make a methodological contribution by developing a selective-control architecture for agentic customer-service operations. The challenge is that stronger controls can prevent failures in operationally coupled requests, but they also impose computation, latency, and customer-facing friction if applied uniformly. Our architecture addresses this challenge by treating control as a routed resource-allocation decision rather than a fixed property of the agent workflow. The difficulty router determines when a session can remain on the baseline path and when operational coupling warrants escalation. For escalated sessions, the workflow applies stronger control at the points where service errors become operationally consequential: when the agent must resolve conflicting instructions and when it is about to commit a state-changing backend write. The evaluation design mirrors this logic by separating operationally conflicted requests from routine requests and examining whether gains reflect structured pre-write control rather than indiscriminate dialogue expansion or broader tool use. Thus, the methodological contribution is a routing-based approach for allocating deliberation, clarification, and write safeguards across heterogeneous service requests.

Third, we make a practical contribution for firms deploying agentic AI in customer-service operations. The findings suggest that operational risk can be managed through control-allocation rules, rather than only through stronger models or broad human oversight. For deployment, the key design principle is to keep routine service on a fast baseline path while routing operationally coupled requests to workflows that clarify ambiguity, preserve customer constraints, and reconsider consequential backend writes before commitment. This gives firms a concrete way to decide where additional control is worth the cost: not on every customer turn, but at decision points where the agent may lose fallback plans, bind records to the wrong action, execute writes in the wrong sequence, or fail to decompose multi-entity requests.

The rest of the paper is organized as follows. Section~\ref{sec:related-work} reviews related research on AI in customer service, conversational tool-agent benchmarks, and control mechanisms for agentic AI. Section~\ref{sec:methodology} presents the difficulty-routed service-control architecture. Section~\ref{sec:experiment-setup} describes the experimental setup. Section~\ref{sec:exp-results-retail} presents the primary empirical analysis in retail service operations. Section~\ref{sec:exp-results-airline_main} presents airline reservation operations as a secondary domain extension for assessing whether the same service-control logic generalizes beyond retail. Section~\ref{sec:conclusion} concludes with implications for deploying agentic AI in customer-service operations.


\section{Related Work}
\label{sec:related-work}

This section situates our work in three related streams: AI in customer service, customer-service benchmarks for conversational tool agents, and control mechanisms for agentic AI. Together, these literatures motivate our focus on difficulty-routed service control, where deliberation, dialogue, and safeguards are allocated selectively across routine and complex customer-service requests.


\subsection{AI in customer service: from consumer response to agentic service execution}

A large body of marketing and service-operations research has examined how customers perceive, accept, and respond to artificial intelligence in service encounters. Foundational service-AI theory classifies service tasks according to the type of intelligence they require---mechanical, analytical, intuitive, and empathetic---and argues that AI can substitute for human labor task by task as its capabilities expand \citep{huang2018artificial}. Related strategic frameworks distinguish mechanical, thinking, and feeling AI and map these capabilities onto marketing research, marketing strategy, and marketing actions \citep{huang2021strategic}. In retailing specifically, AI adoption has been analyzed along dimensions such as whether applications are customer-facing, online or offline, value-creating, and ethically sensitive \citep{guha2021artificial}. This line of work has recently been extended to generative AI in customer care, where interactive GenAI is positioned as a form of feeling AI for emotionally responsive and relationship-oriented service \citep{huang2024caring}. More broadly, the literature has shifted from treating AI as a passive decision aid to studying AI as a semi-autonomous agent. For example, \citet{li2026tools} synthesize evidence on human acceptance of AI as systems shift from tools to semi-autonomous agents capable of decision-making and interaction. Our setting focuses on a more autonomous form of AI service: the service agent is not merely recommending or assisting, but interacting with customers, interpreting complex requests, and executing consequential service actions through backend tools.

A second stream studies whether and when customers accept AI service agents. This literature shows that customer response depends on task characteristics, disclosure, perceived human involvement, and the service context. Algorithm aversion is stronger when tasks are perceived as subjective or requiring human judgment \citep{dietvorst2015algorithm,castelo2019task}. Chatbot aversion can also arise because customers are reluctant to begin with an imperfect first-stage service channel that may require later transfer to a human expert \citep{kagan2026customers}. Disclosure and hybrid human--AI design further shape how customers perceive and communicate with AI service agents \citep{luo2019frontiers,gnewuch2024more}. Together, this literature establishes that the value of AI service agents depends on task completion and on how customers perceive the agent and experience the service interaction \citep{puntoni2021consumers}.

A third stream studies AI as a productive input to service operations. In call-center customer service, voice-based AI can reduce customer complaints, while speech-recognition failures increase customers' demand for human service and complaints \citep{wang2023voice}. Generative-AI assistance can also raise customer-support worker productivity, especially for less-experienced workers \citep{brynjolfsson2025generative}. More recent work pushes beyond assistance toward agentic systems that autonomously perform customer-service tasks, with human-in-the-loop interventions used to manage escalations and recover service quality after AI failures \citep{wang2026agentic}. These studies show that GenAI and agentic AI can reshape service delivery, but they leave open a design problem central to autonomous service operations: once an AI system can both interact with customers and execute backend actions, firms must decide when the agent should proceed routinely, when it should ask clarifying questions, and when it should spend additional reasoning effort before an irreversible write.

Taken together, this literature has largely studied AI as a customer-facing interface, a determinant of customer response, or an assistant that augments worker productivity. Much less attention has been paid to the service-system design problem that arises when an AI agent must both manage customer interaction and execute backend actions. Our paper addresses this gap by studying how firms can allocate deliberation, dialogue, and safeguards across heterogeneous customer-service requests.


\subsection{Customer service as a benchmark for conversational tool agents}

Customer service has long served as a testbed for task-oriented dialogue because it combines natural-language interaction with workflow and policy constraints. Earlier datasets, such as the Action-Based Conversations Dataset \citep{chen2021action}, annotated policy-constrained action sequences in human--human service dialogues. These datasets primarily evaluate whether a model can identify or predict the appropriate action sequence from dialogue. With the rise of large language models and tool-use APIs, customer service has become a natural setting for evaluating interactive agents that must communicate with users, retrieve information, follow business rules, and execute backend actions that change the service state.

The $\tau$-bench environment \citep{yao2024tau} is central to this shift. It places a language agent, equipped with domain-specific APIs and a written policy, in interaction with an LLM-simulated user. Success is evaluated by comparing the final database state and required communications against an annotated task goal. Its retail and airline domains approximate real enterprise service operations, such as returns, exchanges, cancellations, and booking changes, where the agent must elicit information, follow business rules, and update backend systems before committing to the customer. The benchmark also reports pass$^k$, which measures whether the agent succeeds at least once across $k$ repeated attempts on the same task. This shifts attention from whether an agent can produce one fluent transcript to whether it behaves reliably across repeated service interactions.

Subsequent work has made these environments more realistic and demanding. $\tau^{2}$-bench \citep{barres2025tau} introduces a dual-control setting, particularly in telecom support, where both the agent and the user can take actions that affect the shared service state. Related $\tau$-bench auditing and verification efforts show that small ambiguities in policy rules, database targets, or natural-language success criteria can change whether an agent trajectory is judged successful \citep{cuadron2025sabersmallactionsbig}. Other variants stress-test agent robustness by changing the user simulator rather than the service workflow itself. For example, $\tau$-Trait perturbs user traits such as impatience to examine whether agents remain reliable under more challenging customer behavior \citep{he2025impatient}. Beyond evaluation, the same environments have also been used to synthesize agent training data. APIGen-MT \citep{prabhakar2026apigen} produces verified multi-turn trajectories on the $\tau$-bench retail and airline domains through simulated agent-human interplay, using a review committee and execution-based checks to validate task blueprints before collecting trajectories. Notably, it characterizes task difficulty primarily through the number of state-changing ("write") API calls and the policy constraints that couple them—an emphasis consistent with our treatment of backend writes as the locus of operational risk.
Together, these benchmarks establish customer service as a rigorous setting for studying agentic AI, because the agent's output is not only a conversation but also a sequence of consequential operational actions.

Rather than using these benchmarks solely to rank competing agents by aggregate performance, we use them as operational testbeds for a service-design question: how should an agentic system allocate deliberation, clarification, and safeguards across routine and complex requests? In customer service, many requests are straightforward enough for a low-cost baseline flow, but a meaningful share is operationally difficult because it involves interdependent customer goals, conditional fallback instructions, customer confirmations, or state-changing backend actions. These difficult cases create precisely the cost--quality--risk tradeoff faced by firms deploying agentic AI at scale.

\subsection{Service-control mechanisms for agentic AI}

A central design choice for autonomous service agents is how to control their actions before they change backend service records, such as order status, refund decisions, shipping details, or exchange requests. One line of work delegates verification to a \textit{separate} component. LLM-as-a-judge evaluators \citep{zheng2023judging}, generative verifiers \citep{zhang2024generative}, process- and outcome-reward models \citep{lightman2024let}, and agent-as-a-judge schemes \citep{zhuge2024agent} use external models to score, gate, or steer the primary agent. In the $\tau$-bench setting, AGENT-C \citep{kamath2025enforcing} provides a stronger form of runtime control by checking proposed tool calls against formal temporal-policy specifications. When a proposed action would violate the specification, the system blocks that action during generation and forces the agent to produce a compliant alternative. These approaches can improve reliability by placing oversight outside the primary agent, but they also increase the cost and complexity of deploying the service system.

A second line of work lets the agent verify or revise \textit{itself}. Self-verification, self-refinement, and intrinsic self-reflection methods \citep{weng2023large,madaan2023self,li2025inspo} prompt a model to evaluate and revise its own outputs. This design is operationally attractive because it avoids serving a separate verifier, but it is also fragile: a model that judges its own generations can exhibit self-preference bias \citep{panickssery2024llm} and may reproduce the same blind spots that caused the original error. This limitation is especially consequential in service operations, where verification must be targeted rather than universal. Verifying every message would increase cost and latency, whereas skipping verification before a state-changing backend action can permit irreversible operational errors, such as incorrect refunds, cancellations, exchanges, or address changes. Our method follows this self-verification line but introduces write-triggered reconsideration: the agent re-examines its proposed action only when it is about to execute a state-changing backend action. This just-in-time reconsideration concentrates additional reasoning where operational stakes are highest and cost is most justified.

A third stream studies how agents should coordinate reasoning, tool use, and communication during task execution. ReAct \citep{yao2022react} interleaves reasoning traces with tool actions, while Reflexion \citep{shinn2023reflexion} adds verbal self-reflection across attempts. More recent dialogue-oriented templates, such as ReSpAct \citep{dongre2025respact}, make speaking a first-class action, allowing the agent to clarify goals, surface conflicts, and confirm before acting. Yet proactive dialogue also creates an operational tradeoff. Too little clarification leaves ambiguity unresolved and can lead to wrong tool arguments or unsafe writes; too much clarification creates redundant friction, increases latency, fatigues the user, and may distract the agent from the core request \citep{gulati2026ask}. Thus, the value of speaking and reconsidering is \textit{conditional} on the operational risk and structural complexity of the request.

Our method combines these insights in a difficulty-routed service-control architecture for customer-service operations. Rather than applying heavy reasoning, external verification, or proactive dialogue to every customer interaction, the architecture first uses a difficulty router to distinguish routine sessions from structurally complex ones. Routine sessions remain on a low-cost baseline path, while complex sessions are latched to a complex-path control workflow that combines conflict-aware communication, targeted mistake-proofing controls, and write-triggered reconsideration before backend write. This design reflects a central operations tradeoff: high-risk service actions require stronger safeguards, but applying those safeguards indiscriminately would increase cost, latency, and customer friction in routine interactions. Our contribution is to show how agentic AI systems can improve reliability on operationally complex service requests while preserving efficiency for standard service workflows, using retail as the primary empirical setting and airline as a secondary domain for assessing generalizability across customer-service operations.

\section{The Difficulty-Routed Service-Control Architecture}
\label{sec:methodology}

We study customer-service settings in which an autonomous agent interacts with a customer over a full service session. The agent can call backend tools provided by the service system, a setup common in recent tool-using customer-service benchmarks and agentic service settings \citep{yao2024tau,barres2025tau,wang2026agentic}. At each turn, the agent observes the dialogue history and selects the next action. An action may be a customer-facing message, a read-only tool call that retrieves information from firm records, or a state-changing tool call, which we refer to as a \emph{backend write}. Backend writes include issuing a refund, canceling an order, creating a return or exchange, modifying a reservation, or updating shipping information. Because these actions alter firm records and may be costly to reverse, they require stronger control than read-only or communication steps.

\begin{figure}[!htbp]
\centering
\includegraphics[width=0.75\linewidth]{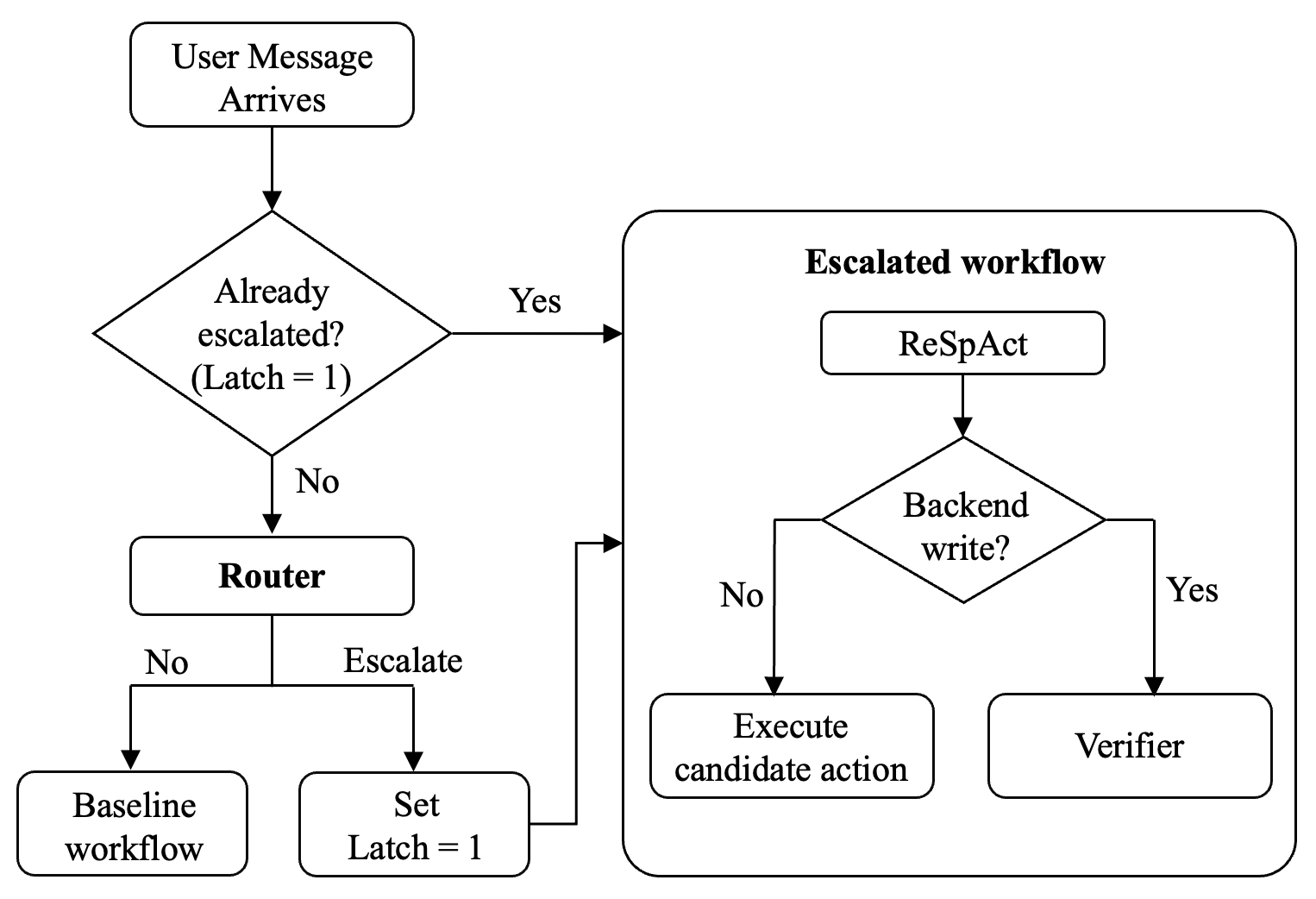}
\caption{Overview of the difficulty-routed service-control architecture. New sessions first pass through a difficulty router. Sessions classified as routine follow the baseline workflow, while sessions classified as operationally coupled are latched to an escalated workflow. Once latched, all subsequent turns bypass the router. The escalated workflow uses ReSpAct candidate generation and applies a pre-write verifier before backend writes.}
\label{fig:method-workflow}
\end{figure}

Our proposed architecture allocates additional control selectively. Routine sessions remain on a baseline path, while sessions with stronger operational coupling are escalated to a higher-control path. The architecture has two components. The first is a \textit{difficulty router}, which decides whether the current session should stay on the baseline path or be escalated. The second is an \textit{escalated workflow}, which governs actions after escalation by combining conflict-aware communication with write-triggered reconsideration through a pre-write verifier before backend writes. Figure~\ref{fig:method-workflow} summarizes the architecture, and the two components are described below.

\subsection{Difficulty Router}

\paragraph{\textbf{Routing Criterion.}}
The router makes the first control decision: whether the session should remain on the baseline path or move to the higher-control escalated path. The routing criterion is \textit{operational coupling}, not surface complexity alone. A session is routed to the escalated path when the dialogue indicates that the agent must coordinate multiple requested service actions, reason across multiple operational entities such as orders, items, reservations, passengers, or payment methods, preserve conditional instructions or fallback requests, resolve conflicting customer constraints, or execute state-changing actions in an order that affects feasibility. For example, a request such as ``exchange it if the requested item is available; otherwise refund me'' requires the agent to preserve both the preferred action and the fallback until feasibility is known. By contrast, a routine cancellation, return, exchange, reservation update, or policy explanation can remain on the baseline path even when the dialogue contains several details. The router is therefore intended as a dispatch rule for control allocation.

\paragraph{\textbf{Formal Routing Rule.}}
Let $M_t$ denote the user message arriving at dialogue turn $t$, and let $S_t$ denote the conversation history after incorporating that message. The system maintains a session-level latch variable $L \in \{0,1\}$, where $L=1$ denotes a session that has already been escalated to the higher-control path and $L=0$ denotes a session that has not yet been escalated.
For each new user turn in a session with $L=0$, the router invokes a router:
\[
    D_t = \Phi(S_t, P), \qquad
    D_t \in \{\texttt{SIMPLE}, \texttt{COMPLEX}\},
\]
where $P$ is a short routing prompt that instructs the model to return only the routing label, either \texttt{SIMPLE} or \texttt{COMPLEX}. The router uses the same language model as the service agent, but the router call is run without tool access. To keep routing lightweight, the call uses a low-cost generation configuration, with a small output budget and minimal reasoning budget when supported by the model provider. 

The latch makes escalation persistent. If $D_t=\texttt{SIMPLE}$, the session remains on the baseline path. If $D_t=\texttt{COMPLEX}$, the system sets $L \leftarrow 1$ and latches the session to the escalated path. Once latched, the router is bypassed for all subsequent turns, and all later agent actions are governed by the escalated workflow. This rule prevents workflow oscillation across turns and ensures that later state-changing actions remain protected by the safeguards triggered earlier in the session.

\paragraph{\textbf{Prompt Implementation.}}
The routing prompts implement this criterion with domain-specific rules. Retail is the primary empirical setting, and airline serves as a structurally distinct secondary domain, so we use separate prompts while keeping the same overall architecture. In retail, operational coupling often appears through cross-order dependencies, conditional refunds or exchanges, item modifications with multiple constraints, and conflicts between the customer's request and retrieved order evidence. The retail prompt therefore emphasizes ambiguity, fallback logic, and multi-object requests. In airline, analogous patterns arise through multi-reservation coordination, multi-segment changes, certificates, payment ordering, and policy-gated changes that require eligibility reasoning. Because airline requests may mention several flights, passengers, or policy restrictions without requiring multiple executable reservation changes, the airline prompt is deliberately more conservative and counts distinct executable operations rather than mentions alone. The prompt templates used in the experiments are shown below.

\vspace{12pt}

\begin{center}
\begingroup
\setlength{\fboxsep}{8pt}
\setlength{\fboxrule}{0.6pt}
\fbox{%
\begin{minipage}{0.91\linewidth}
\small
\noindent\textbf{Prompt Template 1: Retail Difficulty Router}
\vspace{0.4em}

\noindent You are triaging a customer-service request for how much careful planning it needs.

\vspace{0.3em}
\noindent Choose \texttt{COMPLEX} if any of these hold (when unsure, choose \texttt{COMPLEX}):
\begin{itemize}
    \item More than one thing is requested, or more than one order/item is involved.
    \item Any conditional or fallback (``if X is not possible, do Y,'' ``otherwise,'' or a stated preference order).
    \item A return, exchange, cancel, or modify request where the right action is ambiguous, or one action could block another.
    \item The item is lost, damaged, wrong, or the customer is unsure what they want.
    \item Multiple coordinated changes, such as changing both address and items on the same order.
\end{itemize}

\noindent Choose \texttt{SIMPLE} only if it is clearly one straightforward action, with no conditions, no ambiguity, and no interaction between requests.

\vspace{0.3em}
\noindent Answer with exactly one word: \texttt{SIMPLE} or \texttt{COMPLEX}.

\vspace{0.3em}
\noindent Customer conversation so far: \texttt{\{conversation\}}
\end{minipage}%
}
\endgroup
\end{center}

\vspace{12pt}

\begin{center}
\begingroup
\setlength{\fboxsep}{8pt}
\setlength{\fboxrule}{0.6pt}
\fbox{%
\begin{minipage}{0.91\linewidth}
\small
\noindent\textbf{Prompt Template 2: Airline Difficulty Router}
\vspace{0.4em}

\noindent Decide if the customer clearly needs two or more distinct change operations that will actually be carried out on their reservations.

\vspace{0.3em}
\noindent Default to \texttt{SIMPLE}. Only answer \texttt{COMPLEX} when the conversation gives clear evidence of at least two distinct, executable change operations---for example cancelling one reservation and booking or modifying another, changing flights and cabin or baggage on a reservation, or an explicit conditional request.

\vspace{0.3em}
\noindent Answer \texttt{SIMPLE} for:
\begin{itemize}
    \item a single change operation, even if it touches several flights or passengers;
    \item any pure information, price, or flight-status question;
    \item requests the agent will most likely refuse or escalate, such as a cancellation outside business rules, modifying a basic-economy flight, changing the number of passengers, or changing a flight already flown.
\end{itemize}

\noindent Count distinct, executable operations---not flights, not mentions, and not requests that will be declined. When in doubt, answer \texttt{SIMPLE}.

\vspace{0.3em}
\noindent Answer with exactly one word: \texttt{SIMPLE} or \texttt{COMPLEX}.

\vspace{0.3em}
\noindent Customer conversation so far: \texttt{\{conversation\}}
\end{minipage}%
}
\endgroup
\end{center}
\vspace{12pt}

\paragraph{\textbf{Dispatch Procedure.}}
Operationally, the router follows the procedure below:
\begin{enumerate}
    \item Initialize each session with $L=0$.
    \item When a new user message arrives, append it to the dialogue history $S_t$.
    \item If $L=1$, generate the next agent action using the escalated workflow.
    \item If $L=0$, apply the router to $S_t$ and obtain $D_t \in \{\texttt{SIMPLE}, \texttt{COMPLEX}\}$.
    \item If the router returns \texttt{SIMPLE}, keep the session on the baseline path and generate the next agent action using the baseline workflow.
    \item If the router returns \texttt{COMPLEX}, set $L \leftarrow 1$, latch the session to the escalated path, and generate the next and all subsequent agent actions using the escalated workflow.
\end{enumerate}

Escalation adds proactive clarification, targeted safeguards, and a verification step before state-changing actions. These controls can improve reliability on higher-risk interactions, but they also increase computation, latency, and the possibility of extra customer-facing turns. The router therefore allocates the more expensive workflow to structurally coupled requests while preserving the simpler baseline workflow for routine service.

\subsection{Escalated Workflow}

Once a session is latched to the escalated path ($L=1$), the system switches from the baseline workflow to a higher-control workflow. The escalated workflow concentrates additional control at two operationally consequential points. First, it enables \emph{conflict-aware communication}, allowing the agent to clarify ambiguity or explain conflicts before taking action. Second, it applies \emph{write-triggered reconsideration} when the agent is about to execute a backend write. This design reflects a service-operations principle: safeguards are most valuable at decision points where errors are costly or hard to reverse. The first control point is implemented through conflict-aware candidate generation, and the second through pre-write verification.

\paragraph{\textbf{Conflict-Aware Candidate Generation.}}

The escalated workflow begins by generating a candidate next action using a \textit{Reason-Speak-Act} (ReSpAct) prompt \citep{yao2022react,dongre2025respact}. The candidate may be a customer-facing message, a read-only tool call, or a backend write. This prompt treats communication as part of the action space rather than as something that occurs only after tool use fails. When the dialogue contains unresolved conflict---for example, when customer instructions are mutually exclusive, one requested action could block another, or retrieved firm records contradict a customer constraint---the agent can ask a clarification question or explain the conflict before choosing a backend write. In this sense, conflict-aware communication helps the agent preserve customer constraints and resolve ambiguity before a risky state change is attempted.

\paragraph{\textbf{Pre-Write Verification.}}

After a candidate action is generated, the system checks whether it is a backend write. If the candidate is a customer-facing message or a read-only tool call, it proceeds directly because it does not change service records. If the candidate is a backend write, the system pauses before execution and applies a pre-write verification step. We refer to this second-stage check as the \textit{pre-write verifier}. The verifier re-examines the proposed tool name and arguments before commitment. It checks whether the proposed action satisfies the customer request, preserves earlier constraints, uses the correct records and arguments, and avoids blocking later required actions. This step implements write-triggered reconsideration: the agent does not reconsider every turn, but pauses at the write boundary, where an incorrect action would change firm records and may be costly to reverse. The approach is closest to self-verification and self-refinement methods, but applies verification only at the action boundary rather than throughout the dialogue \citep{weng2023large,madaan2023self}.

Backend writes are identified using both domain-specific write sets and benchmark tool metadata. In retail, these tools include order cancellation, return, exchange, and pending-order modification tools. In airline, they include reservation booking, cancellation, certificate, and reservation-update tools. This hybrid rule gives the control layer explicit coverage over known high-risk APIs while preserving compatibility with the domain toolkit.

The resulting escalated workflow turns a single-shot backend write into a two-stage process. The agent first proposes a candidate action under the ReSpAct prompt. If the candidate is a backend write, the pre-write verifier re-evaluates the tool name and arguments before execution. The objective is to keep consequential writes aligned with the customer request, retrieved firm records, and relevant policy constraints. The verifier uses the same language model as the service agent, but with a higher reasoning configuration when supported by the model provider. The method therefore does not require a separate verifier model; it reallocates additional reasoning to the point where checking is most valuable. The prompt template used for pre-write verification is shown below.

\vspace{12pt}

\begin{center}
\begingroup
\setlength{\fboxsep}{8pt}
\setlength{\fboxrule}{0.6pt}
\fbox{%
\begin{minipage}{0.91\linewidth}
\small
\noindent\textbf{Prompt Template 3: Pre-Write Verification}
\vspace{0.4em}

\noindent You are about to perform an \texttt{IRREVERSIBLE} write action. Before committing, think carefully:
\begin{itemize}
    \item Do all of the user's requests still get satisfied if you do this now?
    \item Could this action block a later request, such as returning an order before a later exchange on the same order, cancelling a reservation before a later flight change, or making a change that becomes forbidden after a policy window?
    \item Are the arguments---ids, item/flight/cabin variants, quantities, and passengers---exactly the ones required by the policy and the user's stated constraints?
    \item Does the action preserve earlier constraints that the user has not clearly revoked, especially constraints such as ``do not cancel the whole order''?
\end{itemize}

\noindent If anything is off, do not perform the write: either pick the correct action or send a message to the user. Do not replace a valid write with another confirmation request. Otherwise proceed with the corrected tool call.

\vspace{0.3em}
\noindent Candidate write action(s) to verify before committing:

\noindent\texttt{[\{"name": "...", "arguments": \{...\}\}]}
\end{minipage}%
}
\endgroup
\end{center}
\vspace{12pt}

\paragraph{\textbf{Verifier Outcomes and Mistake-Proofing Controls.}} After verification, the revised output replaces the original candidate. The outcome can take one of four forms: (1) preserve the original action when it is safe and complete; (2) revise the tool call when the original arguments or sequencing are incorrect; (3) ask the user for clarification or confirmation when the constraints cannot be resolved from the current dialogue; or (4) block the action when retrieved firm records contradict the proposed update. The verifier therefore operates as a targeted control layer around risky state changes rather than as a second opinion on every turn.

The implementation also includes localized mistake-proofing controls for recurring service-agent failure modes. A confirmed-write nudge addresses cases where the customer has already approved the proposed action but the agent asks for confirmation again instead of executing the corresponding tool call; in such cases, the system nudges the agent toward the appropriate state-changing tool call. An API-level protocol guard addresses cases where the agent uses the wrong backend tool or supplies information in a format that the tool cannot use to execute the intended update. A retail evidence guard checks the proposed action against the user and order records already retrieved during the session. If the proposed action conflicts with those records, such as canceling the wrong order, modifying the wrong item, or using an ineligible refund method, the guard blocks the action. These controls are designed to be targeted complements to the broader verification step: they address recurring operational failure modes without replacing the general verification layer, in the spirit of recent runtime safeguards for tool-using agents \citep{kamath2025enforcing,cuadron2025sabersmallactionsbig}.

\section{Experiment Setup}
\label{sec:experiment-setup}

We evaluate the difficulty-routed service-control architecture using
human-verified customer-service tasks from $\tau^{2}$-bench
\citep{barres2025tau}. $\tau^{2}$-bench provides
simulated service environments in which an agent interacts with an
LLM-simulated customer, follows written service policies, uses backend tools,
and is evaluated on whether it produces the correct service outcome. Our goal is to measure aggregate task success and study whether stronger
control improves reliability on the service requests where such control should
matter. To focus this analysis, we construct an \emph{evaluation focus set} from
the full benchmark board and characterize its members before reporting routed
agent outcomes.

This section proceeds in three steps. We first describe the benchmark domains, outcome measures, and baseline workflow. We then define the evaluation focus sets used to isolate tasks with operational conflict. Finally, we characterize the focus sets with diagnostic categories that describe the operational control problems posed by their tasks. These diagnostics motivate and interpret the agent runs reported later, but they are not used to train the router, construct prompts, tune run-time interventions, or define results subgroups.

\subsection{Benchmark Domains, Outcome Measures, and Baseline Workflow}

We use two human-verified domains from $\tau^{2}$-bench: retail and airline. The retail domain contains $114$ tasks and represents e-commerce service operations, including returns, exchanges, cancellations, order modifications, payment constraints, inventory availability, and cross-order coordination. We use retail as the primary empirical setting because these operations generate a wide range of operational conflicts at the marketing--operations interface: the agent must preserve customer preferences while executing correct backend writes under policy and inventory constraints. The airline domain contains $50$ tasks and represents reservation-service operations, including booking changes, cancellations, cabin or baggage modifications, multi-segment changes, travel certificates, and policy-gated eligibility constraints. We use airline as a secondary service domain because it provides a different operational structure from retail while still requiring the agent to coordinate customer-facing communication with backend execution. We refer to all tasks available in a domain as its \emph{full task set}.

Each task provides a written service policy, a set of domain tools, an LLM-simulated customer, and an annotated gold solution. The gold solution is one valid way to resolve the request; from it the benchmark derives the evaluation target, namely the backend state the request should produce (e.g., updated order, refund, exchange, or reservation records) together with the information the agent must communicate to the customer. During an episode the agent converses with the customer and reads or updates service records through backend tools to fulfill the request under the policy. Scoring is outcome-based: the resulting backend state must match the target state, and the required communication conditions must be satisfied. Because the criteria are defined over states and communicated information rather than over the gold action sequence, an agent may follow any tool-call trajectory that produces the correct outcome.

Each such run is a episode and receives a binary pass/fail score under the criteria above. Because both the user simulator and the agent are stochastic, the episode-level score for a given task varies across repetitions, so a single run is a noisy estimate of whether a workflow can handle that task. We therefore run each task four times for every workflow and simulator setup and aggregate the four episode-level outcomes into a task-level \emph{majority-pass} rate. A task is counted as a majority pass for a workflow if the workflow succeeds in at least three of its four runs. This aggregation reduces sensitivity to run-level stochastic variation and provides a stable task-level success measure for comparing the baseline workflow with the routed architecture.

The \emph{baseline workflow} is the benchmark's standard service-agent workflow. It uses a single LLM policy that follows the written service policy, interacts with the simulated customer, and executes backend tools directly, without difficulty routing or the escalated workflow proposed in this paper. The baseline uses the same task policies, tool access, user simulator, and outcome scoring as the routed system. We run the baseline on the full task set in each domain to construct the evaluation focus sets described next and to provide the comparison arm for the routed experiments.

\subsection{Evaluation Focus Sets}
\label{sec:evaluation-focus-sets}
The \emph{evaluation focus set} identifies the tasks on which stronger service control is expected to matter. The router is not designed to change every task: routine requests should remain on the baseline path, while the escalated workflow should be used for requests that create \emph{operational conflict}. We define operational conflict as a situation in which completing the request requires the agent to coordinate, revise, or reconcile multiple operational demands, rather than simply execute or decline a single straightforward action. Such conflict may arise from interacting backend writes, incompatible customer goals, changes to a pending action, constraints revealed through firm records, or conditional customer instructions.

We construct the focus set from baseline interactions, before evaluating the routed system. For each task in the full task set, we run the baseline workflow four times using the same service-agent and user-simulator configuration. We then audit each baseline episode transcript using a fixed conflict-labeling prompt. Each transcript includes the customer turns, agent responses, and backend tool calls. The auditing model returns one of two labels: \texttt{CONFLICT} or \texttt{NO\_CONFLICT}. A task enters the evaluation focus set if at least two of its four baseline episodes are labeled \texttt{CONFLICT}. This majority threshold reduces sensitivity to stochastic variation in the simulated dialogue while keeping focus-set membership independent of the routed workflow and its outcomes. Applying the same focus-set construction procedure in both domains gives the retail and airline analyses a common definition of operational conflict, even though retail serves as the primary empirical setting and airline serves as a structurally distinct secondary domain.

The conflict-labeling prompt is provided in the E-companion \ref{ec:prompt-baseline-conflict}. The prompt does not ask for a generic difficulty judgment. Instead, it operationalizes the definition of operational conflict through five specific patterns: \emph{multiple executable writes}, \emph{incompatible intents}, \emph{confirmation-triggered revisions}, \emph{late-emerging constraints}, and \emph{conditional or fallback plans}. For each pattern, the prompt provides a short explanation so that the audit focuses on the operational structure of the episode rather than superficial indicators such as the number of turns or tool calls.

\vspace{-6pt}
Table~\ref{tab:data-stats} reports the focus-set sizes. Retail contains $61$ of $114$ tasks; airline contains $20$ of $50$ tasks.

\begin{table}[!htbp]
\centering
\caption{Verified $\tau^{2}$-bench domains and evaluation focus set.}
\label{tab:data-stats}
\small
\setlength{\tabcolsep}{5pt}
\renewcommand{\arraystretch}{1.15}
\begin{tabular}{@{}l r r@{}}
\toprule
Domain & Full task set & Evaluation focus set \\
\midrule
Retail  & 114 & 61 \\
Airline &  50 & 20 \\
\bottomrule
\end{tabular}

\vspace{0.35em}
\begin{minipage}{\textwidth}
\footnotesize
\textit{Notes.} For each task, we audit the four baseline interaction transcripts with the conflict-labeling prompt. A task enters the focus set if at least two of its four baseline trials are labeled \texttt{CONFLICT}.
\end{minipage}
\end{table}
\vspace{-12pt}

\subsection{Focus-Set Diagnostics}
\label{sec}

The evaluation focus sets identify tasks that exhibit operational conflict, but the conflict label alone does not explain what kind of control problem the agent faces. We therefore further characterize the focus-set tasks to clarify why stronger service control is relevant and how the proposed workflow maps onto the benchmark tasks. Specifically, we manually code the $61$ retail focus-set tasks and the $20$ airline focus-set tasks into domain-specific, non-overlapping diagnostic categories. In the primary retail setting, these diagnostics support the mechanism analysis of routing and verifier behavior. In the secondary airline setting, they assess whether analogous service-control problems arise in a different operational domain. The coding is used only for setup and interpretation: it is not used to train the router, construct prompts, tune run-time interventions, or define results subgroups.

Table~\ref{tab:retail-focus-diagnostics} reports the retail diagnostic distribution and majority-pass performance from the four-trial retail baseline run. In retail, the focus set reflects several recurring service-control demands. Conditional-fallback tasks require the agent to maintain a branch plan until feasibility is resolved, rather than prematurely choosing one action. Confirmation-revision tasks require updating a pending write when the customer changes the requested item, payment method, or cancellation scope. Hidden-evidence tasks require the agent to bind retrieved records, such as addresses, orders, payments, or items, to the correct backend write. Multi-order and multi-write tasks require preserving the association between each order, item, reason, and payment method across several tool calls. Variant, payment, and optimization tasks add another layer of constraint because the agent must search alternatives while preserving price caps, refund preferences, or product variants. Thus, retail difficulty is not simply a matter of long conversations or many tool calls; it arises when the final backend write must preserve a set of linked customer, policy, and record-level constraints.
\vspace{-6pt}

\begin{table}[ht]
\centering
\caption{Retail focus-set diagnostic categories and baseline performance.}
\label{tab:retail-focus-diagnostics}
\small
\setlength{\tabcolsep}{4pt}
\renewcommand{\arraystretch}{1.15}
\begin{tabular}{@{}L{0.27\textwidth} C{0.07\textwidth} C{0.13\textwidth} L{0.40\textwidth}@{}}

\toprule
Diagnostic category & Tasks & Baseline majority pass & Operational control challenge \\
\midrule
Conditional fallback/branch selection
& 17 & 11
& Maintaining if--then preferences, price thresholds, and fallback actions through the final write. \\
Confirmation-triggered request revision
& 10 & 4
& Updating the pending action when the customer revises items, payment method, or cancellation scope after confirmation. \\
Hidden evidence / cross-reference binding
& 11 & 9
& Retrieving missing addresses, orders, payments, or items and binding that evidence to the correct backend action.\\
Multi-order / multi-write orchestration
& 15 & 8
& Coordinating several orders, items, reasons, and write calls without swapping arguments across actions. \\
Variant / payment / optimization constraints
& 8 & 5
& Searching product or payment alternatives while preserving price caps, variant constraints, and refund preferences. \\
\bottomrule
\end{tabular}
\end{table}

\vspace{-6pt}
Table~\ref{tab:airline-focus-diagnostics} reports the airline diagnostic distribution together with majority-pass performance from the four-trial airline baseline run. In airline, the focus set stresses a different mix of service-control problems. Cross-reservation tasks test whether the system escalates when several bookings, passengers, or certificates appear in the request and must not be mixed. Same-reservation two-write tasks require preserving the correct sequence of reservation-level actions, such as applying one change before another cancellation or update. Bundled single-write tasks may appear structurally simple because they involve one backend call, but the call still has to bind the correct passengers, segments, bags, or cabin variants. Payment-ordered tasks are especially difficult because the relevant coupling may be revealed only after the agent retrieves hidden information about certificates, balances, prices, or eligibility. In these cases, the operational difficulty is not always visible from the initial customer request alone.

\begin{table}[ht]
\centering
\caption{Airline focus-set diagnostic categories and baseline performance.}
\label{tab:airline-focus-diagnostics}
\small
\setlength{\tabcolsep}{4pt}
\renewcommand{\arraystretch}{1.15}
\begin{tabular}{@{}L{0.27\textwidth} C{0.07\textwidth} C{0.13\textwidth} L{0.40\textwidth}@{}}
\toprule
Diagnostic category & Tasks & Baseline majority pass & Operational control challenge \\
\midrule
Cross-reservation write orchestration
& 8 & 4
& Coordinating writes across multiple reservations, passengers, or certificates without mixing entity IDs or payment state.\\
Same-reservation two-write coordination
& 5 & 2
& Executing exactly two ordered writes on one reservation, such as an upgrade followed by a cancellation or a reservation change plus baggage update.\\
Bundled multi-item single-write change
& 4 & 3
& Executing one reservation update that bundles several segments, passengers, or bags while preserving variant and cabin constraints. \\
Payment-ordered single-write change
& 3 & 0
& Applying gift-card or certificate balances and price caps \emph{before} the reservation change the customer ultimately requests.\\
\bottomrule
\end{tabular}
\end{table}

\vspace{-6pt}
Together, these diagnostics connect the focus sets to the control logic in Section~\ref{sec:methodology}. They show that focus-set tasks require the agent to preserve linked constraints, bind retrieved records to the correct backend action, or sequence service actions before committing a write. These patterns motivate escalation for operationally coupled requests, conflict-aware communication when customer constraints or firm records conflict, and write-triggered reconsideration before backend writes.
\section{Retail Service Operations: Primary Empirical Analysis}
\label{sec:exp-results-retail}

\definecolor{casebaselinebg}{RGB}{252,245,241}
\definecolor{casebaselineframe}{RGB}{188,101,78}
\definecolor{casebaselineaccent}{RGB}{129,57,39}
\definecolor{casemethodbg}{RGB}{241,247,252}
\definecolor{casemethodframe}{RGB}{72,119,163}
\definecolor{casemethodaccent}{RGB}{36,83,128}
\definecolor{flightbookingnote}{RGB}{34,92,168}
\newcommand{\fbnote}[1]{\textcolor{flightbookingnote}{\textbf{[Flight-booking extension:} #1\textbf{]}}}

This section presents the primary empirical analysis in retail service operations. Retail is the natural setting for the main analysis because it contains a dense set of operational conflicts at the marketing--operations interface, including fallback requests, confirmation-time revisions, cross-order dependencies, inventory and payment constraints, and multi-write execution. The results support three main conclusions. First, the proposed architecture improves reliability most consistently on the evaluation focus set, where operational conflict is expected to matter, rather than uniformly across the full retail task set. Second, the difficulty router allocates escalated control to service sessions with operational conflict, including cases where the conflict is visible in the opening request and cases where it emerges through authentication, record retrieval, policy evidence, or confirmation-stage revision. Third, in stable gain cases, the escalated path helps primarily through write-triggered reconsideration before risky backend actions: it preserves fallback branches, binds retrieved evidence to the correct object and scope, and decomposes or sequences coupled writes before execution. Together, these findings support a conflict-targeted interpretation of the method: the architecture is most useful when customer instructions, firm records, and backend writes must be coordinated before consequential service actions are committed.

\subsection{Full-Task-Set and Focus-Set Performance}

Tables~\ref{tab:retail-fullboard-winrate} and
\ref{tab:retail-focus-winrate} report majority-pass performance at two levels: the full 114-task retail task set and the 61-task retail evaluation focus set. The full task set measures aggregate benchmark performance across all retail tasks, whereas the focus set isolates tasks whose baseline transcripts exhibit operational conflict. This distinction is central to the evaluation design. If difficulty-routed control is a selective service-control policy, its value should appear most clearly on tasks where customer instructions, backend constraints, and execution dependencies are likely to interact, not necessarily on every task in the benchmark.

\vspace{-6pt}
\begin{table}[!htbp]
\centering
\caption{Retail full-task-set majority-pass rate on the 114-task retail task set}
\label{tab:retail-fullboard-winrate}
\small
\setlength{\tabcolsep}{6pt}
\renewcommand{\arraystretch}{1.15}
\begin{tabular}{@{}lcc@{}}
\toprule
Simulator setup & Baseline & Our method \\
\midrule
User Gemini 2.5 / Agent Gemini 3.5 & \texttt{67/114 = 58.8\%} & \texttt{74/114 = 64.9\%} \\
User Gemini 2.5 / Agent ChatGPT 5.5 & \texttt{96/114 = 84.2\%} & \texttt{89/114 = 78.1\%} \\
User Gemini 3.5 / Agent Gemini 3.5 & \texttt{89/114 = 78.1\%} & \texttt{93/114 = 81.6\%} \\
\bottomrule
\end{tabular}
\end{table}

\vspace{-12pt}
\begin{table}[!htbp]
\centering
\caption{Retail focus-set majority-pass rate on the 61-task evaluation focus set.}
\label{tab:retail-focus-winrate}
\small
\setlength{\tabcolsep}{6pt}
\renewcommand{\arraystretch}{1.15}
\begin{tabular}{@{}lcc@{}}
\toprule
Simulator setup & Baseline & Our method \\
\midrule
User Gemini 2.5 / Agent Gemini 3.5 & \texttt{37/61 = 60.7\%} & \texttt{45/61 = 73.8\%} \\
User Gemini 2.5 / Agent ChatGPT 5.5 & \texttt{48/61 = 78.7\%} & \texttt{53/61 = 86.9\%} \\
User Gemini 3.5 / Agent Gemini 3.5 & \texttt{41/61 = 67.2\%} & \texttt{45/61 = 73.8\%} \\
\bottomrule
\end{tabular}
\end{table}

\vspace{-6pt}
These results clarify why the evaluation focus set is the appropriate setting for assessing the value of stronger service control. The focus set is not intended to be uniformly harder across simulator setups; rather, it is designed to isolate operationally conflicted requests. Consistent with this design, baseline majority-pass performance is lower on the focus set than on the full retail task set in two of the three setups, with the largest gap in the User Gemini 3.5 / Agent Gemini 3.5 setup. The focus set therefore provides the sharper test of whether stronger control improves reliability where it is theoretically warranted.

On the evaluation focus set, the routed architecture improves majority-pass performance in all three retail setups. The majority-pass rate increases from $37/61$ to $45/61$ for User Gemini 2.5 / Agent Gemini 3.5, from $48/61$ to $53/61$ for User Gemini 2.5 / Agent ChatGPT 5.5, and from $41/61$ to $45/61$ for User Gemini 3.5 / Agent Gemini 3.5. This consistency is important because the focus set was constructed independently of the routed outcomes and was designed to isolate operationally conflicted service requests. This consistent focus-set improvement supports the selective-control interpretation of the architecture: additional deliberation, clarification, and write-safety checks appear to be especially valuable on retail requests where operational conflict makes routine execution insufficient.

At the full-task-set level, the pattern is more mixed. The routed architecture improves majority-pass performance in two of the three retail setups, increasing from $67/114$ to $74/114$ for User Gemini 2.5 / Agent Gemini 3.5 and from $89/114$ to $93/114$ for User Gemini 3.5 / Agent Gemini 3.5. However, performance declines from $96/114$ to $89/114$ in the User Gemini 2.5 / Agent ChatGPT 5.5 setup. This divergence is informative rather than merely negative. The full task set combines conflict-heavy requests with many routine requests for which the baseline workflow is already sufficient, so aggregate performance can dilute or mask the value of stronger control on the tasks where it is theoretically warranted. The mixed full-task-set pattern therefore reinforces the interpretation of the architecture as a selective service-control policy rather than a universal performance enhancer. The key question is not only whether the method raises average performance over all tasks, but whether it improves reliability on operationally conflicted service sessions and whether the router avoids unnecessary escalation on routine sessions. The next subsection examines whether the router sends the right kinds of sessions to the escalated path.

\subsection{Conflict-Targeted Routing}

The routing results show that \texttt{COMPLEX} activation is concentrated in the evaluation focus set. In the retail results reported above, every task routed into the escalated path belongs to this focus set, which was constructed independently from baseline transcripts before routed outcomes were evaluated. This pattern indicates that the router is identifying service requests with operational conflict rather than simply escalating routine requests.

We examine routing behavior in detail using the primary retail configuration, User Gemini 2.5 / Agent Gemini 3.5, because the other configurations show similar qualitative routing patterns. In this setup, 21 distinct evaluation-focus-set tasks enter the escalated path in at least one run. Because each task is evaluated over repeated stochastic runs, these 21 tasks generate 29 realized escalated-path episodes. We use the task count to describe which types of service requests the router identifies as \texttt{COMPLEX}, and the episode count to examine the conflict patterns that arise during execution. This distinction matters because routing is assigned at the session level, but operational conflict may become visible only through the realized dialogue. Supplementary Table~\ref{tab:ec-retail-routing-summary} in the E-Companion reports the corresponding routing summary.

The routed cases fall into two broad categories. A smaller subset of the realized escalated-path episodes are \emph{immediate-conflict} cases, where the opening request already reveals a non-routine service-control problem. These requests are not simple one-write episodes, such as a single return or cancellation. Instead, the customer may combine multiple actions in the same request, such as cancel plus return, return plus exchange, or order modification plus address change. Other immediate-conflict cases contain explicit fallback or tradeoff structure, such as ``if one option is not feasible, do another,'' or scope and ordering constraints that make an incorrect write especially costly, such as canceling only one item rather than an entire order. In these cases, early escalation reflects operational coupling visible in the customer's first turn.

Most realized escalated-path episodes are \emph{emergent-conflict} cases. In these episodes, the opening request may appear routine, but conflict becomes operationally visible only as the conversation unfolds. These cases arise through several recurring patterns. First, some requests require \emph{retrieval-based target binding}: the customer may ask to use `my default address,'' change `that order,'' or use a preferred payment method, but the agent cannot know the correct address, order, item, or payment record until it retrieves the relevant customer and order information. Second, some requests involve \emph{record-revealed branching}: an exchange, modification, or refund may appear straightforward at first, but inventory, payment, or eligibility evidence later shows that the preferred option is unavailable and a fallback must be considered. Third, some requests involve \emph{pre-write revision}: after the agent has proposed or prepared an action, the customer changes the item, payment method, cancellation scope, or modification preference, so the pending write must be updated before execution. Fourth, some requests involve \emph{order-dependent execution}: especially for pending orders, canceling an order first can change it from a modifiable state to a canceled state, making a later address, item, or payment modification infeasible. These patterns show why routing cannot be reduced to surface features of the opening request; many service-control risks become visible only through interaction with the customer and firm records. E-Companion Section~\ref{ec:retail-routing-conflict} provides additional task-level examples and trigger patterns underlying this immediate-versus-emergent distinction.

The routed episodes also clarify why write-triggered reconsideration is useful after escalation. Once a session enters the escalated path, the agent often faces three recurring execution risks. The first is committing a write to the \emph{wrong object}: the relevant order, item, address, or payment record must be bound from retrieved evidence. The second is committing a write with the \emph{wrong scope}: the requested action may apply to a single item, an entire order, or several coordinated orders, and that scope may shift during the dialogue. The third is committing writes in the \emph{wrong sequence}: especially for pending orders, an early write can change the order state and make a later required update infeasible. These risks map to the retail focus-set diagnostics and provide the operational rationale for checking backend writes before committing them.

Overall, the routing evidence shows that the difficulty router provides the allocation step required by the proposed architecture. It does not merely send linguistically busy sessions to the escalated path; it identifies sessions in which operational conflict is already present or becomes visible through record retrieval, policy or availability constraints, and pre-write revision. This matters because these are the sessions that create the execution risks targeted by the escalated workflow: preserving fallback plans, binding retrieved records to the correct object and scope, and sequencing coupled writes before execution. The routing mechanism therefore helps explain why the method improves most consistently on the evaluation focus set: stronger control is directed toward sessions where routine execution is more likely to be insufficient, while routine out-of-focus requests are not broadly escalated.

\subsection{Sources of Execution Improvement in Escalated Workflow}

To understand where the escalated path changes the execution process, this subsection focuses on \emph{stable gain cases}: retail tasks that enter the escalated path and change from baseline majority failure to routed majority pass. The analysis is descriptive. It uses these cases to characterize how successful escalated-path trajectories differ from baseline failures, rather than to estimate a formal causal decomposition. We focus on the primary retail configuration, User Gemini 2.5 / Agent Gemini 3.5, and use escalated-path episodes from the stable gain cases as the main analysis set. Across these episodes, the recurring pattern is pre-write control: the routed workflow preserves the structure of the customer request, binds retrieved firm records to the correct backend action, and decomposes risky service actions before committing backend writes.

The analysis proceeds in three steps. First, we examine dialogue overhead to assess whether the escalated path creates diffuse customer friction or instead concentrates additional turns around risky write decisions. Second, we examine tool use to assess whether the workflow expands tool use indiscriminately or relies on a limited set of tools for evidence gathering, write separation, and local checking. Third, we compare representative trajectories to describe how the escalated path changes the agent's decision logic before backend writes are committed.

\paragraph{\textbf{Dialogue overhead.}}

The first question is whether escalated-path recovery comes at the cost of diffuse customer-facing friction. Because the escalated workflow can introduce clarification, confirmation, and reconsideration before backend writes, improved performance could simply reflect longer interactions. The relevant issue is therefore whether added turns are generic dialogue overhead or whether they are concentrated around risky write decisions.

Table~\ref{tab:retail-verifier-overhead} reports dialogue statistics for eight \emph{stable gain cases}: Tasks 11, 20, 38, 44, 74, 83, 103, and 104. These cases are useful for this analysis because they are tasks where the escalated path is associated with a change from baseline majority failure to routed majority pass. For each case, we report both the baseline trajectory and the routed trajectory. We report user turns, confirm-like user turns, assistant turns, tool calls, and session duration. \emph{User turns} count customer messages during the service session. \emph{Confirm-like user turns} are customer messages that approve, reject, revise, or finalize a proposed service action before a backend write is executed. \emph{Assistant turns} count non-user agent steps during the session, while \emph{tool calls} separately count backend API calls made by the agent. \emph{Duration} measures elapsed session time in seconds.

Across the eight cases, the routed workflow averages 7.0 user turns, 2.13 confirm-like user turns, 16.5 assistant turns, 9.5 tool calls, and 74.1 seconds, compared with 6.3 user turns, 1.88 confirm-like turns, 15.38 assistant turns, 4.63 tool calls, and 36.3 seconds for the baseline trajectories. The routed workflow therefore does introduce additional operational overhead. However, the table does not support a simple ``longer dialogue'' explanation of the gains. Several successful gain cases remain compact even after routing: Tasks 11, 44, and 83 still involve only 5 to 6 user turns and remain under 50 seconds. Thus, the escalated path does not require turning routine interactions into protracted conversations; rather, it inserts additional control where the request structure makes the write sequence risky.

\begin{table}[!htbp]
\centering
\caption{Case-level dialogue overhead in stable retail gain cases.}
\label{tab:retail-verifier-overhead}
\small
\setlength{\tabcolsep}{5pt}
\renewcommand{\arraystretch}{1.1}
\begin{tabular}{@{}lcccccc@{}}
\toprule
Task & Setting & User turns & Confirm-like turns & Assistant turns & Tool calls & Duration (s) \\
\midrule
\multirow{2}{*}{11} & Baseline & 5 & 1 & 11 & 6 & 22.4 \\
 & Routed & 5 & 1 & 11 & 6 & 38.7 \\
\multirow{2}{*}{20} & Baseline & 4 & 1 & 15 & 10 & 38.9 \\
 & Routed & 5 & 2 & 17 & 12 & 113.4 \\
\multirow{2}{*}{38} & Baseline & 8 & 1 & 19 & 4 & 41.4 \\
 & Routed & 8 & 1 & 19 & 11 & 85.4 \\
\multirow{2}{*}{44} & Baseline & 6 & 2 & 12 & 5 & 22.5 \\
 & Routed & 5 & 1 & 11 & 6 & 39.6 \\
\multirow{2}{*}{74} & Baseline & 5 & 1 & 14 & 2 & 33.8 \\
 & Routed & 8 & 3 & 17 & 9 & 69.3 \\
\multirow{2}{*}{83} & Baseline & 6 & 1 & 13 & 1 & 25.9 \\
 & Routed & 5 & 1 & 12 & 7 & 49.0 \\
\multirow{2}{*}{103} & Baseline & 10 & 5 & 22 & 4 & 63.9 \\
 & Routed & 9 & 4 & 21 & 12 & 86.7 \\
\multirow{2}{*}{104} & Baseline & 6 & 3 & 17 & 5 & 41.6 \\
 & Routed & 11 & 4 & 24 & 13 & 111.0 \\
\midrule
\multirow{2}{*}{Average} & Baseline & 6.3 & 1.88 & 15.38 & 4.63 & 36.3 \\
  & Routed & 7.0 & 2.13 & 16.50 & 9.50 & 74.1 \\
\bottomrule
\end{tabular}
\end{table}

\vspace{-6pt}
The longer gain cases are informative because their added dialogue arises in structurally complex service problems rather than ordinary single-write requests. Tasks 103 and 104 contain 9 and 11 user turns, respectively, with 4 confirmation-like turns in each case. These trajectories involve multi-entity and multi-write requests with several service units, including delivered-order returns and pending-order modifications. In such cases, confirmation turns help partition the request into distinct write blocks before execution.

The distribution of confirmation turns across the gain cases supports this interpretation. Confirmation is not spread uniformly as generic friction; it clusters around specific control points, including fallback resolution, scope- or payment-related reconfirmation, and chunked confirmation of multi-write plans. Task 11 illustrates fallback resolution: the routed trajectory does not add more confirmation turns than baseline, but it spends those turns on settling the policy-compliant fallback rather than prematurely committing the preferred but infeasible refund arrangement. Task 74 illustrates sequencing control: the routed trajectory adds both turns and tool calls because the task contains two distinct service actions, and confirmation is concentrated around the later modification step rather than inserted as general chatter. Task 103 illustrates multi-write chunking: the longer interaction reflects separate confirmation of multiple return and modification blocks rather than uncontrolled workflow verbosity. Thus, the added dialogue is not generic conversational expansion; it is structured pre-write control inserted at operationally risky decision points.

\paragraph{\textbf{Tool-Use Profile.}}

The second question is whether the escalated workflow uses tools in a targeted way or simply uses more tools. Table~\ref{tab:retail-tool-profile} reports tool-use statistics for the baseline and routed trajectories in the same eight stable gain cases. Across these cases, the routed workflow averages 9.5 tool calls, including 2.25 read calls and 2.13 write calls, while using only 3.38 unique tool types on average. The corresponding baseline trajectories average 4.63 total tool calls, 2.25 reads, 2.13 writes, 0.25 generic calls, and 3.38 unique tool types, with a calls-per-user-turn ratio of 0.85. This pattern suggests that the escalated path spends additional tool budget on repeated use of familiar tools rather than by broadening the tool repertoire itself. In operational terms, the workflow appears to use tools for evidence gathering, write separation, and local checking, not for increasingly elaborate tool-chain expansion.

\vspace{-6pt}
\begin{table}[!htbp]
\centering
\caption{Tool-use profile of baseline and routed trajectories in stable retail gain cases.}
\label{tab:retail-tool-profile}
\small
\setlength{\tabcolsep}{4pt}
\renewcommand{\arraystretch}{1.1}
\begin{tabular}{@{}lccccccc@{}}
\toprule
Task & Setting & Total & Read & Write & Generic & Unique tools & Calls / user turn \\
\midrule
\multirow{2}{*}{11} & Baseline & 6 & 4 & 2 & 0 & 4 & 1.20 \\
 & Routed & 6 & 4 & 2 & 0 & 4 & 1.20 \\
\multirow{2}{*}{20} & Baseline & 10 & 9 & 1 & 0 & 5 & 2.50 \\
 & Routed & 12 & 9 & 1 & 0 & 5 & 2.40 \\
\multirow{2}{*}{38} & Baseline & 4 & 2 & 1 & 1 & 4 & 0.50 \\
 & Routed & 11 & 2 & 1 & 1 & 4 & 1.38 \\
\multirow{2}{*}{44} & Baseline & 5 & 3 & 1 & 1 & 5 & 0.83 \\
 & Routed & 6 & 3 & 1 & 1 & 5 & 1.20 \\
\multirow{2}{*}{74} & Baseline & 2 & 0 & 2 & 0 & 2 & 0.40 \\
 & Routed & 9 & 0 & 2 & 0 & 2 & 1.12 \\
\multirow{2}{*}{83} & Baseline & 1 & 0 & 1 & 0 & 1 & 0.17 \\
 & Routed & 7 & 0 & 1 & 0 & 1 & 1.40 \\
\multirow{2}{*}{103} & Baseline & 4 & 0 & 4 & 0 & 3 & 0.40 \\
 & Routed & 12 & 0 & 4 & 0 & 3 & 1.33 \\
\multirow{2}{*}{104} & Baseline & 5 & 0 & 5 & 0 & 3 & 0.83 \\
 & Routed & 13 & 0 & 5 & 0 & 3 & 1.18 \\
\midrule
\multirow{2}{*}{Average} & Baseline & 4.63 & 2.25 & 2.13 & 0.25 & 3.38 & 0.85 \\
 & Routed & 9.50 & 2.25 & 2.13 & 0.25 & 3.38 & 1.40 \\
\bottomrule
\end{tabular}
\end{table}

\vspace{-6pt}
The gain cases separate into two operational profiles. In \emph{retrieval-heavy} cases, such as Tasks 11, 20, and 44, the agent uses read calls to establish user, order, product, or payment evidence before deciding which write is feasible. Task 20 is the clearest example: routed execution increases total calls from 10 to 12 while keeping the same 5 unique tools and the same 9 reads, indicating more persistent evidence use rather than a broader tool repertoire. In \emph{write-coordination-heavy} cases, such as Tasks 74, 103, and 104, tool diversity remains low even though the number of calls is much higher on the routed path. These cases require the agent to split a complex request into separate execution units across orders, items, or stages of the same service request. Task 74 increases from 2 calls to 9 while still using only 2 unique tools, consistent with sequencing rather than tool-chain expansion. Task 103 increases from 4 calls to 12 while remaining at 3 unique tools, reflecting decomposition of a multi-entity request into locally checked write blocks.

This pattern supports the same interpretation suggested by the dialogue-overhead analysis. The escalated path does not appear to recover cases by expanding into a broad set of new tools. Rather, it uses a limited tool set more carefully: retrieving evidence before uncertain writes, separating coupled actions, and executing multi-entity requests as smaller locally verified units.

\paragraph{\textbf{Trajectory-Level Recovery Evidence.}}

We next examine how the routed workflow changes execution trajectories in the stable gain cases. The relevant comparison is not simply whether the routed agent talks more or calls more tools, but whether it avoids the baseline write risks identified above: committing an action to the wrong object, applying the wrong scope, or executing coupled writes in the wrong sequence. The trajectory evidence is consistent with a pre-write control pattern. User feedback can still matter, but typically after the routed workflow has already identified a risky decision point and turned it into a fallback check, scope confirmation, or write-decomposition step. Thus, the main pattern is not that the user supplies substantially new information late in the interaction. Rather, the routed workflow preserves the structure of the original request, binds retrieved evidence to the correct backend action, and decomposes coupled writes before execution.

As shown in the case studies in E-Companion~
\ref{ec:retail-case-studies}, tasks 11, 74, and 103 illustrate three forms of trajectory-level recovery. Task 11 is a case of \emph{conditional-fallback preservation}. The customer wants to return items from two delivered orders and prefers an unusual refund arrangement: refund each order to the other order's payment method. However, the customer also gives a fallback---if that cross-order refund is not allowed, refund each order to its original payment method. The baseline risk is premature commitment: the agent treats the preferred refund arrangement as the execution target and loses the fallback, even though the policy requires the refund to go back to the original payment method for that order. In the routed trajectory, the workflow first recognizes that refunds may need to go back to the payment method associated with each original order, then proceeds with the original-payment fallback once the cross-order option is not supported. This recovery pattern reflects branch preservation and correction rather than materially new late-stage user information.

Task 74 is a case of \emph{write ordering and scope isolation}. The customer wants two actions on different pending orders: first cancel the pending order marked as the priority cancellation, and then modify the separate pending order containing the laptop. The later laptop modification also depends on a payment detail supplied during confirmation: the user specifies the card ending in 2697 for that modification. The baseline risk is that the agent collapses the cancellation and laptop modification into one combined update, instead of keeping them as two ordered actions on different pending orders. This makes it easier to lose the payment preference that applies to the later laptop modification. In the routed trajectory, the workflow completes the cancellation first, then moves to the separate laptop order, preserves the card preference for the modification step, and re-checks the target before the second write. This case shows how the escalated path turns a mixed request into an ordered execution plan, so that the later laptop modification is not overwritten or detached from its payment constraint.

Task 103 is a case of \emph{multi-entity decomposition before writes}. The customer compresses several service units into one request: returning two items from one delivered order, returning another item from a separate delivered order, changing the address on a pending order, modifying an item on that same pending order, and retrieving tracking information. The baseline risk is entity conflation: the agent treats the returns, pending-order updates, and tracking request as one large execution block, so evidence from one order can be applied to the wrong service unit. In the routed trajectory, the workflow handles the request in stages: it first processes the two returns from the same delivered order, then the return from the separate order, then the pending-order address change, then the item modification on that pending order, and finally the tracking request. This case shows how the escalated path preserves entity boundaries by matching each write to its own order, item, and purpose before moving to the next service unit.

Together, these cases show that the escalated path helps primarily through write-triggered reconsideration before risky backend actions. In Task 11, it preserves a fallback branch until the refund constraint has been checked. In Task 74, it keeps two coupled pending-order actions in the correct execution sequence. In Task 103, it decomposes a compressed multi-order request into separate write blocks. Across cases, the routed workflow succeeds by preserving request structure and entity boundaries before committing backend writes.

The trajectory contrasts support the broader interpretation of the escalated path as a pre-write control workflow. The routed workflow adds some dialogue and tool-use overhead, but the case evidence shows that this overhead is concentrated at operationally consequential points: preserving fallback branches, maintaining ordered execution, and separating entity-specific writes before backend actions are committed. The value of the escalated path is therefore not simply greater verbosity or tool intensity, but more structured execution before risky writes.

\section{Airline Reservation Operations: Secondary Empirical Analysis}
\label{sec:exp-results-airline_main}

We use airline reservation operations as a secondary domain to examine whether the service-control logic extends beyond retail. Airline requests create a different form of operational coupling: instead of cross-order returns, refunds, and inventory substitutions, reservation tasks often require payment-instrument ordering, same-day or multi-segment itinerary preservation, policy-gated changes, certificates, and coordinated reservation writes. Thus, the airline setting provides a structurally distinct test of whether difficulty-routed control helps when customer communication must be coordinated with consequential backend reservation updates. We report performance across different airline simulator setups and use the primary configuration (User Gemini 2.5 / Agent Gemini 3.5) for the detailed routing and trajectory evidence summarized below. The other configuration (User Gemini 2.5 / Agent ChatGPT 5.5) shows the same qualitative pattern, so we use the primary configuration for detailed case-level analysis.

The performance results are consistent with the conflict-targeted interpretation from retail. On the full 50-task airline task set, the routed architecture improves majority-pass performance in both simulator setups: from $34/50$ to $39/50$ (68.0\% to 78.0\%) for User Gemini 2.5 / Agent Gemini 3.5, and from $37/50$ to $42/50$ (74.0\% to 84.0\%) for User Gemini 2.5 / Agent ChatGPT 5.5. The gains are larger on the 20-task airline evaluation focus set, where reservation coupling is expected to matter. Focus-set performance increases from $9/20$ to $13/20$ (45.0\% to 65.0\%) in the Gemini-agent setup and from $11/20$ to $13/20$ (55.0\% to 65.0\%) in the ChatGPT-agent setup. These results suggest that stronger control is especially valuable on reservation requests where routine execution is likely to be insufficient, while the full-task-set results show that these focus-set gains do not reduce aggregate airline performance in the completed cells.

Routing evidence supports selective allocation. In the primary airline configuration, all 20 focus-set tasks enter the escalated path in at least one trial. Among routed focus-set sessions, roughly three-fifths are \emph{immediate-conflict} activations, where the opening request exposes cancel-and-rebook structure, cross-reservation coordination, or payment constraints that must be resolved before a reservation change. The remaining two-fifths are \emph{emergent-conflict} activations, where the request initially appears routine but conflict becomes visible only after reservation lookup, flight-search results, policy evidence, or confirmation-stage revision. This split differs from retail, where emergent conflict dominates more strongly, but it reinforces the same point: routing responds to operational coupling rather than merely to surface complexity.

The escalated-path evidence further indicates that airline gains come from structured pre-write control rather than indiscriminate dialogue or tool expansion. In the five stable gain cases, Tasks 15, 16, 21, 30, and 37, additional confirmation and tool calls cluster around operationally consequential points: selecting the correct payment method, rechecking same-day return availability, separating policy-blocked requests from feasible reservation changes, and coordinating flight and baggage writes.

Case-level evidence illustrates these patterns by showing both why the baseline fails and how the escalated path changes execution before the write. Task 16 involves changing a reservation to the cheapest Economy itinerary on the next day, with a refund tied to the original payment method. After the customer confirms the itinerary, the baseline continues discussing whether the refund should go to the gift card or another payment method and never commits the reservation update. The escalated path resolves the payment method before the write, uses the required gift card in the update, and completes the reservation change. In Task 21, the baseline abandons the intended same-day return structure and writes a different return date, while the escalated path preserves the same-day itinerary and coordinates separate flight and baggage writes. Task 30 involves a bundled request with one infeasible part and one feasible part: the customer asks to remove a checked bag and change to a nonstop flight. The baseline explains that bag removal is not allowed but stops after confirmation without executing the feasible flight change. The escalated path separates the policy-blocked baggage request from the feasible nonstop change and commits the latter. These cases show that the same pre-write control logic observed in retail also applies to reservation operations. The detailed airline analysis is reported in E-Companion Section~\ref{ec:exp-results-airline_all}.

\section{Conclusion}
\label{sec:conclusion}

As autonomous agents take on operational execution roles in customer-service systems, firms face a control-allocation problem at the operation--marketing interface. These agents do not only communicate with customers; they retrieve firm records, apply policies, and execute backend writes that change orders, refunds, exchanges, cancellations, reservations, and other service records. Firms must therefore preserve fast, low-friction service for routine requests while preventing costly operational errors when customer instructions, firm records, policy constraints, and backend actions interact.

This paper develops a difficulty-routed service-control architecture for this problem. The central idea is that deliberation, clarification, and safeguards are operational resources that should be allocated selectively. The difficulty router keeps routine sessions on a low-cost baseline path and escalates sessions with operational coupling to a higher-control workflow. Once escalated, the workflow adds control at two points: when the agent must resolve ambiguous or conflicting customer instructions, and when it is about to execute a state-changing backend action. Conflict-aware communication helps surface and resolve competing interpretations before action, while write-triggered reconsideration checks proposed backend writes before commitment. In this way, the architecture concentrates stronger control at high-risk operational decision points rather than applying additional reasoning or confirmation uniformly across all customer turns.

The empirical evidence supports this selective-control interpretation. In the primary retail setting, the method improves performance most consistently on the evaluation focus set, which isolates tasks with operational conflict in baseline transcripts. Aggregate full-task-set effects are more mixed, suggesting that the method should not be interpreted as a universal performance enhancer. Instead, its value lies in allocating stronger control to the cases where additional safeguards are most likely to matter. The routing analysis shows that escalated-path activation is concentrated on conflict-heavy tasks rather than routine requests. Dialogue and tool-use profiles suggest that gains do not come from indiscriminate interaction expansion or broader tool chains; instead, added turns and tool calls support evidence gathering, write separation, local checking, and pre-write reconsideration. Case-level evidence indicates that gains arise from better pre-write control: preserving fallback plans, binding retrieved records to the correct backend action, sequencing coupled writes, and decomposing multi-entity requests into locally checked execution units. Airline reservation operations provide a structurally distinct secondary domain for examining whether the same service-control logic extends beyond retail.

The paper contributes to research on agentic AI in customer-service operations in three ways. First, it reframes autonomous service agents as service-control systems at the operations--marketing interface, where customer communication must be coordinated with backend writes that determine the operational outcome of the service request. Second, it develops a routing-based architecture for selective control allocation. The architecture treats clarification, deliberation, and write safeguards as controls to be allocated across requests rather than as fixed features of a single agent workflow. Routine sessions remain on a low-cost baseline path, while operationally coupled sessions receive stronger pre-write control before consequential backend actions. Third, it provides an evaluation approach for studying whether stronger control is allocated to the requests that need it. By distinguishing conflict-heavy service requests from routine requests and analyzing performance, routing, dialogue overhead, tool use, and trajectory-level evidence, this approach complements aggregate benchmark evaluation with a service-control question: whether the agent applies the right level of control to the right type of service request.

The results also have practical implications for firms deploying agentic AI. Stronger models and human oversight are important, but they are not the only levers for managing operational risk. Firms can also design routing and write-safety policies that determine when an agent should proceed routinely, when it should ask clarifying questions, and when it should reconsider a proposed backend write before committing it. Such policies can help preserve customer experience on routine interactions while reducing the likelihood of incorrect refunds, cancellations, exchanges, or reservation changes on operationally complex requests.

Several limitations point to future research. First, the present study evaluates control allocation in benchmarked service environments, where task goals, service policies, tools, and final states are specified. Future work could study difficulty-routed control in live service settings, where customer patience, trust, channel switching, and repeat-contact behavior affect the cost and value of clarification or reconsideration. Second, our analysis focuses on task success, routing behavior, dialogue overhead, tool use, and trajectory-level evidence. Future research could develop richer cost models that jointly account for reliability, latency, customer friction, tool-use cost, and the expected cost of incorrect backend writes. Third, the current architecture uses prompt-based routing and write-triggered reconsideration. Future work could compare the prompt-based router and pre-write verifier with learned routing models, explicit policy-checking systems, or hybrid human--AI escalation rules, especially in domains where regulatory, financial, or safety consequences make backend-write errors especially costly.

Overall, this paper shows that the reliability of agentic customer-service systems is not only a modeling problem, but also an allocation problem. As autonomous agents take on more operational responsibility, firms will need to decide where speed should be preserved, where ambiguity should be surfaced, and where safeguards should be inserted before action. Difficulty-routed service control offers one approach to making these choices explicit. Service agents should \emph{reconsider} when customer instructions, firm records, policy constraints, and backend writes become operationally coupled.

\bibliography{ref1}

\newpage
\definecolor{casebaselinebg}{RGB}{252,245,241}
\definecolor{casebaselineframe}{RGB}{188,101,78}
\definecolor{casebaselineaccent}{RGB}{129,57,39}
\definecolor{casemethodbg}{RGB}{241,247,252}
\definecolor{casemethodframe}{RGB}{72,119,163}
\definecolor{casemethodaccent}{RGB}{36,83,128}
\definecolor{flightbookingnote}{RGB}{34,92,168}

\ECSwitch
\ECHead{E-Companion}
\label{sec:e_companion}


\section{Prompt Template for Baseline Conflict Labeling}
\label{ec:prompt-baseline-conflict}

\vspace{12pt}
\begin{center}
\begingroup
\setlength{\fboxsep}{8pt}
\setlength{\fboxrule}{0.6pt}
\fbox{%
\begin{minipage}{0.91\linewidth}
\small
\noindent\textbf{Prompt Template: Baseline Conflict Labeling}
\vspace{0.4em}

\noindent You are auditing \texttt{ONE} customer-service episode for \texttt{OPERATIONAL CONFLICT}.

\noindent You are given the full transcript of a single baseline episode: the customer's turns, the agent's replies, and every backend tool call with its result (\texttt{USER:}, \texttt{AGENT:}, \texttt{TOOL:} lines). Decide whether correctly completing this customer's request required resolving operational conflict---i.e., it was not a single straightforward action but involved interacting, competing, or multiply-coordinated operational demands.

\noindent Label \texttt{CONFLICT} if the episode shows \texttt{ANY} of:
\begin{itemize}
    \item \textbf{Multiple executable writes:} the request requires two or more distinct backend changes---whether competing \emph{or} merely needing coordination so their targets and arguments (ids, items, reasons, payment) are not mixed up.
    \item \textbf{Incompatible intents:} the customer asks for things that cannot all be satisfied under policy, forcing a trade-off or the refusal of one part.
    \item \textbf{Confirmation-triggered revision:} the customer changes the requested action (item, scope, payment, option) after the agent has proposed or confirmed it, so a pending action must be revised before it executes.
    \item \textbf{Late-emerging constraint:} a relevant restriction, price, balance, or eligibility surfaces only after the agent retrieves backend records, changing what can be done.
    \item \textbf{Conditional / fallback plan:} an explicit ``if X is not possible, then do Y'' the agent must carry until feasibility is resolved.
\end{itemize}

\noindent Label \texttt{NO\_CONFLICT} if the request is a single straightforward action (even if it touches several items or takes several turns), a pure information / status / price question, or a request that is simply fulfilled or simply declined without any of the above. A single backend write is never, by itself, a conflict.

\noindent Judge what the episode actually required---not the number of turns or tool calls. When the evidence is unclear, default to \texttt{NO\_CONFLICT}.

\vspace{0.3em}
\noindent Output exactly one token: \texttt{CONFLICT} or \texttt{NO\_CONFLICT}.

\vspace{0.3em}
\noindent Transcript:

\noindent\texttt{\{transcript\}}
\end{minipage}%
}
\endgroup
\end{center}
\vspace{12pt}

\newpage
\section{Routing Summary for the Primary Retail Configuration}
\label{ec:routing-summary}
This section reports supplementary routing statistics for the primary retail configuration, User Gemini 2.5 / Agent Gemini 3.5. The table summarizes the number of retail evaluation focus-set tasks routed to the \texttt{COMPLEX} path, the corresponding realized escalated-path episodes across repeated trials, and the main trigger families observed in those routed cases.

\begin{table}[ht]
\centering
\caption{Routing summary for the primary retail configuration (User Gemini 2.5 / Agent Gemini 3.5).}
\label{tab:ec-retail-routing-summary}
\small
\setlength{\tabcolsep}{6pt}
\renewcommand{\arraystretch}{1.15}
\begin{tabular}{@{}p{0.36\textwidth}p{0.18\textwidth}p{0.34\textwidth}@{}}
\toprule
Quantity & Value & Interpretation \\
\midrule
Retail focus-class tasks & 61 & Conflict-prone retail subset used for routing analysis \\
Distinct tasks routed to \texttt{COMPLEX} & 21 & Task-level classifier outcome: 21 focus-class tasks are assigned to the escalated path \\
Realized escalated-path episodes & 29 & Run-level execution outcome: repeated trials over the 21 tasks produce 29 escalated episodes \\
Out-of-focus-class \texttt{COMPLEX} routings & 0 & No evidence of unnecessary escalation on routine retail tasks \\
Dominant trigger families & fallback / branch selection; multi-write coordination; retrieval- or confirmation-revealed conflict & Operational conflict, not surface complexity, is the main routing signal \\
\bottomrule
\end{tabular}
\end{table}

\newpage
\section{Immediate and Emergent Conflict in Retail Routing}
\label{ec:retail-routing-conflict}

This section provides additional detail on the retail routing patterns summarized in the main text. The main results distinguish between two types of \texttt{COMPLEX} activation. In \emph{immediate-conflict} cases, the customer's opening request already reveals operational coupling, so the router can escalate early. In \emph{emergent-conflict} cases, the request initially appears routine, but the need for stronger control becomes visible only after authentication, record retrieval, policy or availability evidence, or confirmation-stage revision. The examples below describe the main trigger patterns in each group and show why the router is responding to operational conflict rather than surface linguistic complexity.

\paragraph{Immediate conflict at the opening turn.}
In the first category, the user's initial request already exposes clear
operational coupling, so the router has little reason to wait for additional
tool evidence before escalating to the \texttt{COMPLEX} path. One common trigger
is \emph{multiple coupled operations in the same opening request}. These are not
simple one-write episodes such as a single return or a single cancellation, but
requests that immediately combine actions such as cancel plus return, return
plus exchange, or modify plus address change. Representative retail tasks in
this category include 16, 27, 41, 98, and 99. A second trigger is an
\emph{explicit fallback or tradeoff structure} in the opening turn. Here the
user says, in effect, ``if one option is not feasible, do another,'' or imposes
cost-saving or scope-limiting preferences that turn the episode into a
branch-selection problem rather than a straightforward write. Representative
tasks include 19, 28, and 35. A third trigger is that the opening request
contains \emph{scope or ordering constraints} that make an incorrect write
especially costly, for example when the user distinguishes cancelling a single
item from cancelling an entire order, or when a required action order is stated
up front. Task 28 is representative. Finally, some openings explicitly involve
\emph{multiple orders or multiple write targets} from the start, as in Tasks 16,
28, 98, and 99. Across these cases, the key point is that the user's own first
turn already reveals that the episode is not a one-step routine action. The
router is therefore responding to operational coupling, not to superficial
linguistic complexity.

\paragraph{Emergent conflict during the dialogue.}
The second category is equally important because it captures the kinds of cases
for which routing and later verifier activation are most valuable. In these
episodes, the opening request may look routine, but conflict becomes
operationalized only as the conversation unfolds. One common pathway is that
\emph{authentication and retrieval are needed before the write target can be
bound correctly}. The user may refer to ``my default address'' or ``that order''
in a way that sounds simple, yet the actual write target becomes identifiable
only after profile or order evidence is retrieved. Representative tasks include
42, 72, 96, and 104. A second pathway is that \emph{availability or policy
evidence turns an apparent routine exchange into a genuine fallback branch}. For
example, a preferred item variant may be unavailable, forcing the system to
preserve and resolve an ``if not A, then B'' structure after tool evidence
appears. Representative tasks include 1, 30, and 31. A third pathway is
\emph{confirmation-stage scope revision}. This is a particularly important class
because the user changes payment method, narrows the modification scope,
withdraws part of the request, or turns a whole-order modification into a
single-item change at the moment just before a write would occur. Representative
tasks include 71, 72, and 98. A fourth pathway is that \emph{multi-write order
dependence becomes visible only after tool constraints are known}. For pending
orders, one modification may lock out another, so what initially appears to be a
normal multi-step request becomes a sequence-sensitive service-control problem.
Representative tasks include 41, 96, 103, and 104. Finally, \emph{partial versus
whole-order ambiguity} and \emph{payment reassignment or refund-channel changes}
often emerge only after clarification and tool feedback, as in Tasks 28, 76,
81, 11, 71, 72, and 98. These cases show that the routed path is not only for
requests that are obviously complex at the start; it is also for requests whose
conflict structure is progressively revealed through interaction. Across these
routed retail episodes, three trigger families appear especially important:
\emph{fallback and branch selection}, \emph{multi-write or multi-order
coordination}, and \emph{retrieval- or confirmation-revealed conflict}. The
last of these includes hidden target binding, confirmation-time scope revision,
payment reassignment, and same-order write-order dependence, and appears to be
the dominant emergent-conflict pattern in the current retail sample.

\newpage
\section{Retail Case Trajectory Excerpts}
\label{ec:retail-case-studies}

\begin{figure}[!htbp]
\centering
\begin{minipage}[t]{0.475\textwidth}
\fcolorbox{casebaselineframe}{casebaselinebg}{%
\begin{minipage}[t]{0.94\linewidth}
\colorbox{casebaselineframe}{\parbox{\dimexpr\linewidth-2\fboxsep\relax}{\textbf{Task 11: Baseline Failure}}}\\[4pt]
\textbf{User intent:} refund two delivered orders; prefer cross-order refund routing, but accept refund to the original payment method if the preferred routing is not allowed.\\[2pt]
\textbf{Representative trajectory excerpt:}
\begin{itemize}
\item \textbf{Baseline agent:} ``I will process the returns and try to refund each order to the other order's payment method.''
\item \textbf{Baseline agent:} ``The orders can be returned, so I'll proceed with the requested refund handling.''
\item \textbf{Baseline agent:} ``Your return has been processed.''
\end{itemize}
\textbf{\textcolor{casebaselineaccent}{Failure point:}} The baseline treats the user's first preference as the direct execution target and does not preserve the ``otherwise refund to the original payment method'' fallback as an active branch. In effect, it commits too early, before the policy boundary on cross-order refund routing is fully resolved.
\end{minipage}}
\end{minipage}
\hfill
\begin{minipage}[t]{0.475\textwidth}
\fcolorbox{casemethodframe}{casemethodbg}{%
\begin{minipage}[t]{0.94\linewidth}
\colorbox{casemethodframe}{\parbox{\dimexpr\linewidth-2\fboxsep\relax}{\textbf{Task 11: Routed Recovery}}}\\[4pt]
\textbf{Routing status:} \texttt{COMPLEX} with verifier-style control before the write.\\[2pt]
\textbf{Representative trajectory excerpt:}
\begin{itemize}
\item \textbf{Method agent:} ``I can help process the returns, but the refund must go back through the payment method associated with each original order.''
\item \textbf{Method agent:} ``If refunding to the other order's payment method is not supported, I will proceed with the original payment methods instead.''
\item \textbf{Method agent:} ``The returns have now been submitted under the original payment methods for the two orders.''
\end{itemize}
\textbf{\textcolor{casemethodaccent}{Why this case matters:}} The method succeeds by preserving the user's fallback structure until the policy constraint is resolved, rather than collapsing the conditional request into a single invalid refund plan. The gain comes from branch preservation and correction, not from new late-stage user information.
\end{minipage}}
\end{minipage}
\caption{Case study 1: Task 11. The routed method succeeds by preserving the fallback structure until the policy constraint is resolved.}
\label{fig:case11}
\end{figure}

\begin{figure}[!htbp]
\centering
\begin{minipage}[t]{0.475\textwidth}
\fcolorbox{casebaselineframe}{casebaselinebg}{%
\begin{minipage}[t]{0.94\linewidth}
\colorbox{casebaselineframe}{\parbox{\dimexpr\linewidth-2\fboxsep\relax}{\textbf{Task 74: Baseline Failure}}}\\[4pt]
\textbf{User intent:} complete two coupled actions across different pending orders, with a confirmation-stage preference that matters for the later write.\\[2pt]
\textbf{Representative trajectory excerpt:}
\begin{itemize}
\item \textbf{Baseline agent:} ``I can take care of both requests for you.''
\item \textbf{Baseline agent:} ``I'll update the laptop order and cancel the other pending order as requested.''
\item \textbf{Baseline agent:} ``The changes have been applied.''
\end{itemize}
\textbf{\textcolor{casebaselineaccent}{Failure point:}} The baseline does not fail because it misunderstands the request at the language level. It fails because it does not separate the two risky writes into independent control units. The user-specified execution order is lost, and the two orders are handled as an under-controlled combined action sequence.
\end{minipage}}
\end{minipage}
\hfill
\begin{minipage}[t]{0.475\textwidth}
\fcolorbox{casemethodframe}{casemethodbg}{%
\begin{minipage}[t]{0.94\linewidth}
\colorbox{casemethodframe}{\parbox{\dimexpr\linewidth-2\fboxsep\relax}{\textbf{Task 74: Routed Recovery}}}\\[4pt]
\textbf{Routing status:} \texttt{COMPLEX} with targeted confirmation before the second write.\\[2pt]
\textbf{Representative trajectory excerpt:}
\begin{itemize}
\item \textbf{Method agent:} ``I will first cancel the pending order you marked as the priority cancellation.''
\item \textbf{Method agent:} ``That cancellation is complete. I'll now move to the separate pending order containing the laptop.''
\item \textbf{Method agent:} ``Before I submit the modification, I'll use the specified card ending in 2697 if confirmation is needed.''
\item \textbf{Method agent:} ``The laptop modification has now been applied to the second order.''
\end{itemize}
\textbf{\textcolor{casemethodaccent}{Why this case matters:}} The method separates two coupled writes and respects the user-specified execution order, whereas the baseline collapses them into a single under-controlled action sequence. The gain comes from turning a sequence constraint into execution structure.
\end{minipage}}
\end{minipage}
\caption{Case study 2: Task 74. The routed method improves by separating two mutually constraining writes and re-checking scope before the second write.}
\label{fig:case74}
\end{figure}

\begin{figure}[!htbp]
\centering
\begin{minipage}[t]{0.475\textwidth}
\fcolorbox{casebaselineframe}{casebaselinebg}{%
\begin{minipage}[t]{0.94\linewidth}
\colorbox{casebaselineframe}{\parbox{\dimexpr\linewidth-2\fboxsep\relax}{\textbf{Task 103: Baseline Failure}}}\\[4pt]
\textbf{User intent:} combine several service units in one dialogue, including returns, pending-order modification, address change, and a tracking-related request.\\[2pt]
\textbf{Representative trajectory excerpt:}
\begin{itemize}
\item \textbf{Baseline agent:} ``I can handle the returns, update the pending order, and get the tracking information for you.''
\item \textbf{Baseline agent:} ``I'll process the returned items and make the address and item changes together.''
\item \textbf{Baseline agent:} ``Everything has been updated.''
\end{itemize}
\textbf{\textcolor{casebaselineaccent}{Failure point:}} This is a classic entity-conflation failure. The baseline treats several service units as one large execution block, so evidence from different delivered and pending orders is insufficiently separated before write execution. The problem is not missing tools; it is loss of local consistency across entities.
\end{minipage}}
\end{minipage}
\hfill
\begin{minipage}[t]{0.475\textwidth}
\fcolorbox{casemethodframe}{casemethodbg}{%
\begin{minipage}[t]{0.94\linewidth}
\colorbox{casemethodframe}{\parbox{\dimexpr\linewidth-2\fboxsep\relax}{\textbf{Task 103: Routed Recovery}}}\\[4pt]
\textbf{Routing status:} \texttt{COMPLEX} with local verification before each write block.\\[2pt]
\textbf{Representative trajectory excerpt:}
\begin{itemize}
\item \textbf{Method agent:} ``I will handle these in stages: first the two returns from the same delivered order, then the return from the separate order.''
\item \textbf{Method agent:} ``Those returns are complete. I'll now update the pending order address.''
\item \textbf{Method agent:} ``The address change is done. Next I'll modify the item on that same pending order.''
\item \textbf{Method agent:} ``I can now retrieve the tracking details separately.''
\end{itemize}
\textbf{\textcolor{casemethodaccent}{Why this case matters:}} The method decomposes a multi-entity request into locally verified write units, preventing evidence from one order from leaking into another. The gain comes from local evidence binding before each write rather than from additional generic deliberation.
\end{minipage}}
\end{minipage}
\caption{Case study 3: Task 103. The routed method recovers by decomposing a multi-entity request into locally verified write units.}
\label{fig:case103}
\end{figure}

\newpage
\section{Immediate and Emergent Conflict in Airline Routing}
\label{ec:airline-routing-conflict}

This section provides additional detail on the airline routing patterns
summarized in Section~\ref{sec:airline-routing}. The analysis is restricted to
the airline evaluation focus set ($20$ tasks) in the primary configuration,
User Gemini~2.5 / Agent Gemini~3.5. The main results distinguish between two
types of \texttt{COMPLEX} activation. In \emph{immediate-conflict} cases, the
customer's opening request already reveals reservation-level operational
coupling, so the router can escalate on the first user turn. In
\emph{emergent-conflict} cases, the request initially appears routine, but the
need for stronger control becomes visible only after authentication, reservation
retrieval, flight-search feasibility, policy evidence, or confirmation-stage
revision. Across the $80$ focus-class episodes, $49$ activate on the opening
turn and $31$ activate only after dialogue unfolds. The examples below describe
the main trigger patterns in each group and show why the router is responding
to operational conflict rather than surface linguistic complexity.

\paragraph{Immediate conflict at the opening turn.}
In the first category, the user's initial request already exposes clear
operational coupling, so the router has little reason to wait for additional
tool evidence before escalating to the \texttt{COMPLEX} escalated-path. One common
trigger is \emph{cross-reservation or multi-booking coordination} in the
opening turn. These are not simple one-write reservation updates, but requests
that immediately involve several bookings, passengers, or certificates.
Representative focus tasks include 7, 14, 15, 33, 39, and~44. A second trigger
is \emph{cancel-and-rebook or cancel-all structure} stated up front: the
customer asks to cancel one or more reservations and then pursue a replacement
itinerary or rebooking path rather than a single in-place modification.
Representative tasks include 23 and~29. A third trigger is an explicit
\emph{payment-before-change} cue in the opening request, where gift-card
balances, certificates, or refund-channel preferences make payment ordering
part of the service problem from the start. Representative tasks include 16
and~37. A fourth trigger is a \emph{bundled multi-segment or multi-passenger
change} where an incorrect first write would be especially costly because
several segments, bags, or passengers must move together. Representative tasks
include 8, 11, and~40. Across these cases, the key point is that the user's
own first turn already reveals that the episode is not a one-step routine
reservation action. The router is therefore responding to operational coupling,
not to superficial linguistic complexity.

\paragraph{Emergent conflict during the dialogue.}
The second category is equally important because it captures the kinds of cases
for which routing and later escalated-path control are most valuable. In these
episodes, the opening request may look routine, but conflict becomes
operationalized only as the conversation unfolds. One common pathway is that
\emph{reservation retrieval and flight search are needed before the feasible
write can be bound}. The user may request a schedule change that sounds
straightforward, yet same-day return feasibility, basic-economy constraints,
or segment availability become knowable only after lookup and search.
Representative tasks include 8, 17, 21, 22, and~42. A second pathway is that
\emph{policy or eligibility evidence turns an apparent routine modification
into a decomposed request}. For example, a bundled customer request may mix a
feasible flight change with a policy-blocked baggage action, and the feasible
sub-request must be isolated only after the policy gate is stated. Task~30 is
representative. A third pathway is \emph{confirmation-stage payment or scope
revision}. The user clarifies gift-card preference, refund channel, or baggage
handling just before a write would occur, turning a seemingly settled plan into
a payment-ordering problem. Representative tasks include 11, 16, 30, and~32.
A fourth pathway is \emph{same-reservation multi-write dependence}: a flight
update and a separate baggage update must remain distinct control units with
coordinated but separate payment handling, and that structure becomes clear only
after the agent proposes concrete itinerary options. Representative tasks
include 21 and~32. Finally, \emph{conditional cancel-or-rebook fallback}
structure may be present in the task design but only become executable after
the agent discovers that an in-place modification path is blocked---for example
by a basic-economy segment---and must preserve the fallback branch before
writing. Task~21 is representative. These cases show that the routed path is
not only for requests that are complex at the start; it is also for
requests whose conflict structure is progressively revealed through interaction
with reservation records and search tools. Across these routed airline episodes,
three trigger families appear especially important: \emph{cancel-and-rebook and
cross-reservation coordination}, \emph{payment-instrument ordering before
writes}, and \emph{retrieval- or confirmation-revealed conflict}. The last of
these includes flight-search feasibility, policy-gated decomposition,
confirmation-time payment preference, and same-reservation write sequencing, and
accounts for the minority of focus tasks that never latch on the opening turn
in any trial (Tasks~8, 17, 21, 22, 30, 32, and~42).

\newpage

\section{Additional Analysis Results of Airline Reservation Operations}
\label{ec:exp-results-airline_all}


This section presents the complementary empirical analysis in flight-booking
service operations. Airline complements retail as a smaller but structurally
distinct stress test: failures concentrate on reservation write orchestration,
including cancel-and-rebook fallbacks, payment-instrument ordering, bundled
segment updates, and cross-reservation coordination, rather than on the
cross-order refund routing that dominates retail. The results support three
main conclusions. First, the proposed architecture improves reliability most
clearly on the airline evaluation focus set, where reservation coupling is
expected to matter, while also yielding full-board gains in both
simulator setups rather than diluting performance across routine tasks.
Second, on the focus set, conflict-targeted escalation identifies sessions with
genuine reservation-control risk, including cases where coupling is visible in
the opening request and cases where it emerges through reservation retrieval,
flight-search feasibility, policy evidence, or confirmation-stage revision.
Third, in stable gain cases, the escalated path helps primarily through
write-triggered reconsideration before risky reservation actions: it binds
payment instruments, preserves same-day return and bundled-change structure,
and decomposes or sequences coupled flight and baggage writes before execution.
Together, these findings support the same conflict-targeted interpretation as
in retail: the architecture is most useful when customer instructions,
reservation records, and backend writes must be coordinated before consequential
service actions are committed.

\subsection{Full-Task-Set and Focus-Set Performance}
\label{sec:airline-full-focus}

Tables~\ref{tab:airline-fullboard-winrate} and
\ref{tab:airline-focus-winrate} report majority-pass performance at two
levels: the full $50$-task airline task set and the $20$-task airline
evaluation focus set. The full task set measures aggregate benchmark
performance across all verified airline tasks, whereas the focus set isolates
tasks whose gold solutions require reservation write orchestration under
operational coupling. This distinction
parallels the retail evaluation design: if difficulty-routed control is
valuable, its effect should appear most clearly where payment ordering,
cancel-and-rebook fallbacks, and multi-write coordination are expected to
matter, not only in aggregate performance over the full board.

\begin{table}[!htbp]
\centering
\caption{Airline full-task-set majority-pass rate on the 50-task airline task set.}
\label{tab:airline-fullboard-winrate}
\small
\setlength{\tabcolsep}{6pt}
\renewcommand{\arraystretch}{1.15}
\begin{tabular}{@{}lcc@{}}
\toprule
Simulator setup & Baseline & Our method \\
\midrule
User Gemini 2.5 / Agent Gemini 3.5 & \texttt{34/50 = 68.0\%} & \texttt{39/50 = 78.0\%} \\
User Gemini 2.5 / Agent ChatGPT 5.5 & \texttt{37/50 = 74.0\%} & \texttt{42/50 = 84.0\%} \\
\bottomrule
\end{tabular}
\end{table}

\begin{table}[!htbp]
\centering
\caption{Airline focus-set majority-pass rate on the 20-task evaluation focus set.}
\label{tab:airline-focus-winrate}
\small
\setlength{\tabcolsep}{6pt}
\renewcommand{\arraystretch}{1.15}
\begin{tabular}{@{}lcc@{}}
\toprule
Simulator setup & Baseline & Our method \\
\midrule
User Gemini 2.5 / Agent Gemini 3.5 & \texttt{9/20 = 45.0\%} & \texttt{13/20 = 65.0\%} \\
User Gemini 2.5 / Agent ChatGPT 5.5 & \texttt{11/20 = 55.0\%} & \texttt{13/20 = 65.0\%} \\
\bottomrule
\end{tabular}
\end{table}

These results clarify why the airline evaluation focus set is the appropriate
setting for assessing stronger reservation control. In both simulator
setups, baseline majority-pass performance is lower on the focus set than on
the full task set. For User Gemini~2.5 / Agent Gemini~3.5, baseline performance
falls from $34/50$ to $9/20$; for User Gemini~2.5 / Agent ChatGPT~5.5, from
$37/50$ to $11/20$. This pattern is consistent with the construction of the
focus set: it isolates structurally conflicted reservation episodes rather than
a random subset of easier tasks. The focus set therefore provides the sharper
test of whether difficulty-routed control improves reliability where routine
execution is more likely to be insufficient.

On the evaluation focus set, the routed architecture improves majority-pass
performance in both setups. Performance increases from $9/20$ to
$13/20$ for User Gemini~2.5 / Agent Gemini~3.5 and from $11/20$ to $13/20$
for User Gemini~2.5 / Agent ChatGPT~5.5. Although the tied focus-set headline
rate ($65.0\%$) masks different baseline difficulty across agent families, both
cells show positive focus-set movement relative to their own matched baselines.
This is important because the focus set was defined from task structure and gold
solutions before routed outcomes were evaluated. The consistent focus-set
improvement supports the interpretation that additional deliberation,
confirmation, and write-safety control are especially valuable on airline requests
where reservation coupling makes routine execution insufficient.

At the full-task-set level, both rows show the same directional
pattern rather than the mixed outcome seen in retail. Majority-pass performance
increases from $34/50$ to $39/50$ for User Gemini~2.5 / Agent Gemini~3.5 and
from $37/50$ to $42/50$ for User Gemini~2.5 / Agent ChatGPT~5.5. The full-board
gains are not driven by the focus subset alone: in the Gemini~3.5-agent cell,
non-focus tasks move from $25/30$ to $26/30$ under majority pass, while in the
GPT-5.5-agent cell they move from $26/30$ to $29/30$. Together with the larger
focus-set gains, this suggests that stronger control helps on structurally
conflicted reservation tasks while still yielding some benefit on the remaining
board tasks.

Taken together, the airline performance evidence in Tables~\ref{tab:airline-fullboard-winrate}
and~\ref{tab:airline-focus-winrate} supports the same evaluative logic as
retail, with a smaller but structurally distinct board. The focus set remains
the primary setting for mechanism interpretation; the full-board rows show that
focus-set gains are not an artifact of evaluating only the hardest tasks. The
next subsection examines whether escalated sessions on the focus set correspond
to immediate and emergent reservation conflict rather than indiscriminate
routing (Section~\ref{sec:airline-routing}).

\subsection{Conflict-Targeted Routing}
\label{sec:airline-routing}

We analyze conflict-targeted routing on the airline evaluation focus set
($20$ tasks), defined independently of routed outcomes. In the primary configuration,
User Gemini~2.5 / Agent Gemini~3.5, all $20$ focus tasks enter the
\texttt{COMPLEX} path in at least one trial; there are no task-level
\texttt{SIMPLE} assignments on this set. 

In this setup, each focus task is evaluated over four stochastic trials. We use
the task count to describe which reservation-request types are escalated, and
within-task trial variation to describe when conflict becomes visible during
execution. This distinction matters because routing is a session-level
decision, but payment constraints, feasibility evidence, and write risk often
surface only after reservation retrieval or flight search.

Among routed focus-class sessions, escalation timing falls into two broad
categories. Roughly three-fifths are \emph{immediate-conflict} activations: the
opening customer request already exposes cancel-and-rebook structure,
cross-reservation coordination, or payment-before-change cues, and escalation
occurs on the first user turn. About two-fifths are \emph{emergent-conflict}
activations: the opening request appears routine, but conflict becomes visible
only after reservation lookup, flight-search feasibility, policy evidence, or
confirmation-stage revision. This pattern differs from retail, where
immediate-conflict activations are a smaller minority; on airline focus tasks,
early observable triggers account for a larger share of first-turn escalation,
but a substantial minority of routed sessions still enter the escalated path only
once retrieved evidence makes the write risk operational. E-Companion
Section~\ref{ec:airline-routing-conflict} provides additional task-level
examples and trigger patterns underlying this immediate-versus-emergent
distinction.

Once a session enters the escalated path, the agent faces three recurring
execution risks analogous to retail. The first is writing with the \emph{wrong
payment instrument or refund channel}: gift-card balances, certificates, and
original-payment preferences must be bound from retrieved evidence before a
reservation write. The second is writing the \emph{wrong reservation
structure}: same-day returns, bundled segments, and passenger sets must be
preserved through confirmation. The third is \emph{failing to execute a
feasible reservation write after dialogue completion}: the agent may resolve a
policy gate in conversation but stop before committing the reservation update.
These risks map to the airline focus-set diagnostics
(Table~\ref{tab:airline-focus-diagnostics}) and motivate write-triggered
reconsideration before irreversible reservation actions.

Taken together, the airline routing evidence shows that conflict-targeted
escalation is concentrated on the evaluation focus set, where reservation
write orchestration is expected to matter. Within this set, the router
identifies operationally coupled requests rather than escalating at random, as
shown by the split between immediate- and emergent-conflict activation. The
recovery analysis below examines whether the escalated workflow improves
pre-write control on the focus tasks where baseline majority failure changes
to routed majority pass (Section~\ref{sec:airline-verifier}).

\subsection{Sources of Execution Improvement in Escalated Workflow}
\label{sec:airline-verifier}

This subsection examines recovery patterns in airline cases where the
routed workflow changes a baseline majority failure into a routed majority pass.
The evidence is descriptive, but it points to the same pre-write control pattern
as in retail: successful routed trajectories preserve the structure of the
customer request, gather or bind the relevant reservation and payment records,
and decompose risky service actions before committing backend writes. We focus on
the primary airline configuration, User Gemini~2.5 / Agent Gemini~3.5,
and use escalated-path episodes from the airline \emph{stable gain cases} as the
main analysis set. In the focus-class evaluation, five tasks meet this criterion:
Tasks~15, 16, 21, 30, and~37.

The analysis proceeds in three steps. First, we examine whether the workflow adds
excessive dialogue overhead or whether additional turns are concentrated around
risky write decisions. Second, we examine whether the workflow expands tool use
indiscriminately or instead uses a limited set of tools for evidence gathering
and write decomposition. Third, we compare representative trajectories to
describe how the escalated path changes decision logic before the write boundary.

\paragraph{\textbf{Dialogue overhead.}}

The first concern is customer friction. Because the escalated workflow can
introduce additional clarification, confirmation, and reconsideration before
irreversible reservation writes, performance gains may come at the cost of longer
interactions. The relevant question is therefore whether the additional turns
are diffuse conversational overhead or whether they are concentrated around
high-risk write decisions.

Table~\ref{tab:airline-verifier-overhead} reports dialogue statistics for the
five airline stable gain cases. For each case, we report both the baseline
trajectory and the routed trajectory. Across the five cases, the routed
workflow averages 6.2 user turns, 2.2 confirm-like user turns, 14.6 assistant
turns, 8.4 tool calls, and 335.8 seconds, compared with 7.6 user turns, 2.6
confirm-like turns, 16.4 assistant turns, 8.8 tool calls, and 416.4 seconds for
the baseline trajectories. The routed workflow therefore does not uniformly
lengthen customer-facing dialogue; on average, routed episodes are slightly more
compact in user turns and total duration. However, the table does not support a
simple ``shorter dialogue'' explanation of the gains either. Task~30 remains
relatively compact on the routed path---six user turns, two confirm-like turns,
four tool calls, and 278.9 seconds---despite involving a bundled outbound flight
change with a policy gate on baggage removal. Thus, escalated workflow does
not require turning routine interactions into protracted conversations.

\begin{table}[!htbp]
\centering
\caption{Case-level dialogue overhead in stable airline gain cases.}
\label{tab:airline-verifier-overhead}
\small
\setlength{\tabcolsep}{5pt}
\renewcommand{\arraystretch}{1.1}
\begin{tabular}{@{}lcccccc@{}}
\toprule
Task & Setting & User turns & Confirm-like turns & Assistant turns & Tool calls & Duration (s) \\
\midrule
\multirow{2}{*}{15} & Baseline & 11 & 2 & 19 & 8 & 349.2 \\
 & Routed & 6 & 2 & 14 & 8 & 361.8 \\
\multirow{2}{*}{16} & Baseline & 6 & 2 & 12 & 6 & 155.0 \\
 & Routed & 6 & 2 & 14 & 8 & 372.3 \\
\multirow{2}{*}{21} & Baseline & 9 & 3 & 23 & 14 & 994.2 \\
 & Routed & 7 & 2 & 19 & 12 & 391.4 \\
\multirow{2}{*}{30} & Baseline & 6 & 3 & 10 & 4 & 172.4 \\
 & Routed & 6 & 2 & 10 & 4 & 278.9 \\
\multirow{2}{*}{37} & Baseline & 6 & 3 & 18 & 12 & 411.2 \\
 & Routed & 6 & 3 & 16 & 10 & 274.6 \\
\midrule
\multirow{2}{*}{Average} & Baseline & 7.6 & 2.6 & 16.4 & 8.8 & 416.4 \\
 & Routed & 6.2 & 2.2 & 14.6 & 8.4 & 335.8 \\
\bottomrule
\end{tabular}
\end{table}

The longer gain cases are informative because they involve structurally complex
reservation problems rather than ordinary single-write requests. Task~21 is the
clearest example: the baseline trajectory contains 9 user turns, 3 confirm-like
turns, and 994.2 seconds, whereas the routed trajectory compresses the episode
to 7 user turns and 391.4 seconds while still executing coordinated flight and
baggage writes. Tasks~16 and~37 require flight search, payment-instrument
selection, and one or two reservation writes; their routed trajectories add
assistant and tool activity around decomposition of the service problem into
search, payment confirmation, and write execution rather than around generic
workflow verbosity.

The distribution of confirmation turns across the gain cases supports this
interpretation. Confirmation is not spread uniformly as generic friction; it
clusters around specific control points, including payment-instrument binding,
same-day return feasibility, policy-gated bundled changes, and certificate
coordination. Task~16 illustrates payment-ordered control: after the user
confirms the itinerary, the baseline trajectory re-opens refund routing, whereas
the successful routed trajectory uses confirmation to settle payment on the
required gift card before the reservation write. Task~21 illustrates sequencing
control: the episode contains distinct flight and baggage writes, and
confirmation is concentrated around binding the same-day return structure rather
than inserted as general chatter. Task~30 illustrates policy-gated bundled
control: the agent states the baggage-removal constraint, isolates the feasible
flight-change request, and confirms the nonstop option before executing a single
reservation update. Together, these cases show that the escalated path adds
targeted control turns at the points where they are operationally useful.

\paragraph{\textbf{Tool-Use Profile.}}

A second concern is whether the escalated workflow uses tools in a targeted
way or simply uses more tools. Table~\ref{tab:airline-tool-profile} reports
tool-use statistics for the baseline and routed trajectories in the same five
stable gain cases. Across these cases, the routed workflow averages 8.4 tool
calls, including 6.6 read calls and 1.2 write calls, while using 5.6 unique
tool types on average. The corresponding baseline trajectories average 8.8 total
tool calls, 6.8 reads, 1.0 writes, 1.0 generic calls, and 5.4 unique tool types,
with a calls-per-user-turn ratio of 1.19; the routed trajectories average 1.34
calls per user turn. This pattern suggests that the escalatedpath spends
additional tool budget on repeated use of familiar airline tools rather than by
broadening the tool repertoire itself. In operational terms, the workflow appears
to use tools for reservation and user lookup, flight search, write separation,
and local checking---not for increasingly elaborate tool-chain expansion.

\begin{table}[!htbp]
\centering
\caption{Tool-use profile of baseline and routed trajectories in stable airline gain cases.}
\label{tab:airline-tool-profile}
\small
\setlength{\tabcolsep}{4pt}
\renewcommand{\arraystretch}{1.1}
\begin{tabular}{@{}lccccccc@{}}
\toprule
Task & Setting & Total & Read & Write & Generic & Unique tools & Calls / user turn \\
\midrule
\multirow{2}{*}{15} & Baseline & 8 & 7 & 1 & 0 & 5 & 0.73 \\
 & Routed & 8 & 7 & 1 & 0 & 5 & 1.33 \\
\multirow{2}{*}{16} & Baseline & 6 & 5 & 0 & 1 & 5 & 1.00 \\
 & Routed & 8 & 7 & 1 & 0 & 6 & 1.33 \\
\multirow{2}{*}{21} & Baseline & 14 & 12 & 2 & 0 & 6 & 1.56 \\
 & Routed & 12 & 9 & 2 & 1 & 7 & 1.71 \\
\multirow{2}{*}{30} & Baseline & 4 & 3 & 1 & 0 & 4 & 0.67 \\
 & Routed & 4 & 3 & 1 & 0 & 4 & 0.67 \\
\multirow{2}{*}{37} & Baseline & 12 & 7 & 1 & 4 & 7 & 2.00 \\
 & Routed & 10 & 7 & 1 & 2 & 6 & 1.67 \\
\midrule
\multirow{2}{*}{Average}& Baseline & 8.8 & 6.8 & 1.0 & 1.0 & 5.4 & 1.19 \\
& Routed & 8.4 & 6.6 & 1.2 & 0.6 & 5.6 & 1.34 \\
\bottomrule
\end{tabular}
\end{table}

The gain cases separate into two operational profiles. In \emph{retrieval-heavy}
cases, such as Tasks~15, 16, and~37, the agent uses read calls to establish
reservation, user, and flight evidence before deciding which write is feasible and
which payment instrument to bind. Task~16 is representative: routed execution
increases total calls from 6 to 8 while expanding reads from 5 to 7 before a
single \texttt{update\_reservation\_flights} write. In \emph{write-coordination-heavy}
cases, such as Task~21, tool diversity remains modest even though the number of
writes is higher. Task~21 uses 12 routed calls but only 7 unique tools, with
separate flight and baggage updates, consistent with sequencing rather than
tool-chain expansion. Task~30 occupies a compact third pattern: four total tool
calls and four unique tools on both baseline and routed paths, showing that
escalated-path control can remain light when the episode compresses into one
well-checked reservation update after a policy gate.

This pattern supports the same mechanism suggested by the dialogue-overhead
analysis. The escalated path does not appear to recover cases by wandering into
increasingly elaborate tool workflows. Rather, it uses a limited tool set more
carefully: retrieving evidence before uncertain writes, separating coupled
actions, and executing multi-part reservation requests as smaller locally
verified units.

\paragraph{\textbf{Trajectory-Level Recovery Evidence.}}

We next examine how the routed workflow changes execution trajectories in the
stable gain cases. The relevant comparison is not simply whether the routed
agent talks more or calls more tools, but whether it avoids the baseline write
risks identified above: committing an action with the wrong payment instrument,
applying the wrong return-date structure, or stopping after confirmation without
executing a feasible reservation write.

The trajectory evidence is consistent with a pre-write control pattern. User
feedback can still matter, but typically after the routed workflow has already
identified a risky decision point and turned it into a payment check, scope
confirmation, or write-decomposition step. Thus, the main pattern is not that the
user supplies substantially new information late in the interaction. Rather, the
routed workflow preserves the structure of the original request, binds retrieved
evidence to the correct backend action, and decomposes coupled writes before
execution.

As shown in the trajectory excerpts in E-Companion~
\ref{ec:airline-case-studies}, Tasks~16, 21, and~30 illustrate three forms of
trajectory-level recovery. Task~16 is a case of \emph{payment-ordered write
control}. The customer requests a cabin downgrade to the cheapest Economy
itinerary on the next day and accepts refund to the original payment method. The
gold write requires applying the correct gift-card payment instrument on
\texttt{update\_reservation\_flights}. The baseline risk is stalling at the
payment/refund boundary: after the user confirms the itinerary, the baseline
trajectory re-opens refund routing and never executes the write with the
required payment argument. In the routed trajectory, the workflow resolves the
payment instrument explicitly and completes the reservation update.

Task~21 is a case of \emph{same-day return binding and multi-write sequencing}.
The customer wants the fastest return on the same day as the May~27 departure
and also wants baggage added with gift-card payment rules respected. The
baseline risk is to accept a May~28 return after mis-estimating same-day
feasibility, then execute flight and baggage writes that no longer match the
intended same-day structure. In the routed trajectory, the workflow re-checks
same-day availability after retrieval, binds the May~27 return segments, and
executes the flight update and baggage update as separate but coordinated
writes.

Task~30 is a case of \emph{policy-gated bundled-change control}. The customer
requests a nonstop outbound change and also asks to remove a checked bag. The
baseline risk is to resolve both requests in dialogue but stop after user
confirmation without executing the feasible flight change, leaving the database
unchanged despite an apparently successful dialogue ending. In the routed
trajectory, the workflow first states the baggage-removal policy gate, isolates
the feasible flight-change request, confirms the nonstop option and refund
channel, and then executes \texttt{update\_reservation\_flights}.

Taken together, these cases show that the escalated path helps primarily through
write-triggered reconsideration before risky backend actions. In Task~16, it
binds the correct payment instrument before the reservation update. In Task~21,
it preserves same-day return structure across coordinated flight and baggage
writes. In Task~30, it separates a policy-blocked baggage request from a
feasible bundled flight change and completes the latter. Across the three cases,
the routed workflow succeeds by preserving request structure and reservation
boundaries before committing backend writes.

The trajectory contrasts support the broader interpretation of the escalated path
as a pre-write control workflow. The routed workflow adds some dialogue and
tool-use overhead, but the case evidence shows that this overhead is
concentrated at operationally consequential points: binding payment instruments,
preserving same-day return structure, and separating policy-blocked requests
from feasible reservation writes before backend actions are committed. The value
of the escalated path is therefore not simply greater verbosity or tool intensity,
but more structured execution before risky writes.

\section{Airline Case Trajectory Excerpts}
\label{ec:airline-case-studies}

This section provides the trajectory excerpts summarized in
Section~\ref{sec:airline-verifier}. The cases are drawn from the primary airline
configuration, User Gemini~2.5 / Agent Gemini~3.5. Figures~\ref{fig:airline-case16}--\ref{fig:airline-case30}
illustrate stable gain cases.

\begin{figure}[ht]
\centering
\begin{minipage}[t]{0.475\textwidth}
\fcolorbox{casebaselineframe}{casebaselinebg}{%
\begin{minipage}[t]{0.94\linewidth}
\raggedright\sloppy\small
\colorbox{casebaselineframe}{\parbox{\dimexpr\linewidth-2\fboxsep\relax}{\textbf{Task 16: Baseline Failure}}}\\[4pt]
\textbf{User intent:} change reservation M05KNL to the cheapest Economy
itinerary on the next day; accept refund to the original payment method.\\[2pt]
\textbf{Representative trajectory excerpt:}
\begin{itemize}
\item \textbf{Baseline agent:} ``Please reply with `yes' or `confirm' to authorize the update \dots cabin class to Economy \dots''
\item \textbf{User:} ``Yes, I confirm the flight changes. However, for the refund, I'd like it to go back to the original payment method \dots not the gift card.''
\item \textbf{Baseline agent:} ``The original payment method \dots was indeed the Gift Card \dots the refund of \$2,580 will be credited back \dots''
\end{itemize}
\textbf{\textcolor{casebaselineaccent}{Failure point:}} The baseline re-opens refund routing after confirmation and never executes \texttt{update\_reservation\_flights} with the required gift-card payment argument.
\end{minipage}}
\end{minipage}
\hfill
\begin{minipage}[t]{0.475\textwidth}
\fcolorbox{casemethodframe}{casemethodbg}{%
\begin{minipage}[t]{0.94\linewidth}
\raggedright\sloppy\small
\colorbox{casemethodframe}{\parbox{\dimexpr\linewidth-2\fboxsep\relax}{\textbf{Task 16: Routed Recovery}}}\\[4pt]
\textbf{Routing status:} \texttt{COMPLEX} with payment-ordered reconsideration before the write.\\[2pt]
\textbf{Representative trajectory excerpt:}
\begin{itemize}
\item \textbf{Method agent:} ``Option 1 (Cheapest) \dots Total Price: \$207 \dots Refund to you: \$2,580 \dots to your gift card ending in 8887175.''
\item \textbf{User:} ``I'll go with Option 1 \dots I'm happy with the refund being processed to the gift card.''
\item \textbf{Method agent:} Calls \texttt{update\_reservation\_flights} with payment on gift card \texttt{8887175}; reservation update succeeds.
\end{itemize}
\textbf{\textcolor{casemethodaccent}{Why this case matters:}} The method binds the correct payment instrument and completes the reservation update, whereas the baseline stalls at the refund/payment boundary.
\end{minipage}}
\end{minipage}
\caption{Case study 1: Task~16. The routed method succeeds by resolving payment-ordered write control before the reservation update.}
\label{fig:airline-case16}
\end{figure}

\begin{figure}[ht]
\centering
\begin{minipage}[t]{0.475\textwidth}
\fcolorbox{casebaselineframe}{casebaselinebg}{%
\begin{minipage}[t]{0.94\linewidth}
\raggedright\sloppy\small
\colorbox{casebaselineframe}{\parbox{\dimexpr\linewidth-2\fboxsep\relax}{\textbf{Task 21: Baseline Failure}}}\\[4pt]
\textbf{User intent:} change the return on reservation OBUT9V to the fastest
same-day option on May~27; add one checked bag; use appropriate gift-card
payment.\\[2pt]
\textbf{Representative trajectory excerpt:}
\begin{itemize}
\item \textbf{Baseline agent:} ``Your outbound flight \dots arrives in Denver at 19:00 \dots returning on May~27th isn't possible.''
\item \textbf{User:} ``Okay \dots find the fastest return option for me on May~28th instead.''
\item \textbf{Baseline agent:} Calls \texttt{update\_reservation\_flights} and \texttt{update\_reservation\_baggages} for a May~28 return.
\end{itemize}
\textbf{\textcolor{casebaselineaccent}{Failure point:}} The baseline abandons the same-day return structure and executes writes on the wrong return date, producing a database mismatch even though the dialogue appears complete.
\end{minipage}}
\end{minipage}
\hfill
\begin{minipage}[t]{0.475\textwidth}
\fcolorbox{casemethodframe}{casemethodbg}{%
\begin{minipage}[t]{0.94\linewidth}
\raggedright\sloppy\small
\colorbox{casemethodframe}{\parbox{\dimexpr\linewidth-2\fboxsep\relax}{\textbf{Task 21: Routed Recovery}}}\\[4pt]
\textbf{Routing status:} \texttt{COMPLEX} with coordinated flight-and-baggage writes.\\[2pt]
\textbf{Representative trajectory excerpt:}
\begin{itemize}
\item \textbf{Method agent:} ``After you arrive in Denver at 6:00 AM on May~27th, the fastest available return option \dots HAT290 \dots HAT175 \dots''
\item \textbf{User:} ``Yes, please go ahead \dots add one more checked bag \dots use the gift card with the smallest balance.''
\item \textbf{Method agent:} Calls \texttt{update\_reservation\_flights} and \texttt{update\_reservation\_baggages} for the May~27 return.
\end{itemize}
\textbf{\textcolor{casemethodaccent}{Why this case matters:}} The method preserves the same-day return plan and executes the flight and baggage writes as separate but coordinated control units.
\end{minipage}}
\end{minipage}
\caption{Case study 2: Task~21. The routed method improves by binding the same-day return structure before coordinated flight and baggage writes.}
\label{fig:airline-case21}
\end{figure}

\begin{figure}[ht]
\centering
\begin{minipage}[t]{0.475\textwidth}
\fcolorbox{casebaselineframe}{casebaselinebg}{%
\begin{minipage}[t]{0.94\linewidth}
\raggedright\sloppy\small
\colorbox{casebaselineframe}{\parbox{\dimexpr\linewidth-2\fboxsep\relax}{\textbf{Task 30: Baseline Failure}}}\\[4pt]
\textbf{User intent:} change the LAS--IAH outbound from one-stop to nonstop
(HAT266); also request checked-bag removal and refund if possible.\\[2pt]
\textbf{Representative trajectory excerpt:}
\begin{itemize}
\item \textbf{Baseline agent:} ``Checked bags cannot be removed or refunded after the initial booking \dots For the nonstop flight \dots HAT266 \dots''
\item \textbf{User:} ``I'd like to go with flight HAT266 \dots please use the gift card for payment.''
\item \textbf{User:} ``Yes, I confirm these changes and would like to proceed.''
\end{itemize}
\textbf{\textcolor{casebaselineaccent}{Failure point:}} The baseline resolves the policy gate in dialogue but stops after user confirmation without executing \texttt{update\_reservation\_flights}.
\end{minipage}}
\end{minipage}
\hfill
\begin{minipage}[t]{0.475\textwidth}
\fcolorbox{casemethodframe}{casemethodbg}{%
\begin{minipage}[t]{0.94\linewidth}
\raggedright\sloppy\small
\colorbox{casemethodframe}{\parbox{\dimexpr\linewidth-2\fboxsep\relax}{\textbf{Task 30: Routed Recovery}}}\\[4pt]
\textbf{Routing status:} \texttt{COMPLEX} with policy-gated bundled-change control.\\[2pt]
\textbf{Representative trajectory excerpt:}
\begin{itemize}
\item \textbf{Method agent:} ``We cannot remove the bag or refund the bag fee \dots I can help you change your outbound one-stop to nonstop HAT266.''
\item \textbf{User:} ``Yes, please proceed with the flight change to HAT266 \dots send the refund to the original payment method.''
\item \textbf{Method agent:} Calls \texttt{update\_reservation\_flights} for nonstop HAT266.
\item \textbf{Method agent:} ``Your outbound flight is now nonstop HAT266 \dots refund of \$414 \dots''
\end{itemize}
\textbf{\textcolor{casemethodaccent}{Why this case matters:}} The method isolates the feasible flight-change request under a policy gate and completes the bundled reservation update that the baseline leaves unexecuted.
\end{minipage}}
\end{minipage}
\caption{Case study 3: Task~30. The routed method recovers by completing the feasible bundled flight change after a policy-blocked baggage request.}
\label{fig:airline-case30}
\end{figure}

\end{document}